\documentclass[sn-mathphys,openbib]{sn-jnl}

\usepackage[utf8]{inputenc}

\usepackage{hyperref}       
\usepackage{nicefrac}       
\usepackage{microtype}      
\usepackage{comment}
\usepackage{mathtools}
\usepackage{wrapfig}
\usepackage[font=small]{caption}
\usepackage{subcaption}
\usepackage[english]{babel}
\usepackage{bm}
\usepackage{url}
\usepackage{empheq}
\usepackage{array}
\usepackage{nameref}

\jyear{2021}%

\theoremstyle{thmstyleone}%
\theoremstyle{thmstyletwo}%
\newtheorem{example}{Example}%
\newtheorem{remark}{Remark}%

\theoremstyle{thmstylethree}%

\def\eqref#1{equation~(\ref{#1})}
\def\Eqref#1{Equation~(\ref{#1})}

\begin{document}

\title[Goal-Oriented Sensitivity Analysis of Hyperparameters in Deep Learning]{Goal-Oriented Sensitivity Analysis of Hyperparameters in Deep Learning}


\author*[1,2,3]{\fnm{Paul} \sur{Novello}}\email{paul.novello@outlook.fr}

\author[1]{\fnm{Gaël} \sur{Poëtte}}\email{gael.poette@cea.fr}

\author[2]{\fnm{David} \sur{Lugato}}\email{david.lugato@cea.fr}

\author[2,3]{\fnm{Pietro Marco} \sur{Congedo}}\email{pietro.congedo@inria.fr}

\affil[1]{\orgdiv{CESTA}, \orgname{CEA}, \orgaddress{\city{Le Barp}, \postcode{33114}, \country{France}}}

\affil[2]{\orgdiv{CMAP}, \orgname{Ecole Polytechnique}, \orgaddress{\city{Palaiseau}, \postcode{91120}, \country{France}}}

\affil[3]{\orgdiv{Platon}, \orgname{Inria Paris Saclay}, \orgaddress{\city{Palaiseau}, \postcode{91120}, \country{France}}}


\def\rvx{{\mathbf{x}}}
\def\rvy{{\mathbf{y}}}
\def\rvepsilon{{\boldsymbol{\epsilon}}}
\def\rvtheta{{\boldsymbol{\vtheta}}}
\def\rvz{{\mathbf{z}}}

\def\vmu{{\boldsymbol{\mu}}}
\def\vtheta{{\boldsymbol{\theta}}}
\def\valpha{{\boldsymbol{\alpha}}}
\def\vepsilon{{\boldsymbol{\epsilon}}}
\def\vsigma{{\boldsymbol{\sigma}}}
\def\valpha{{\boldsymbol{\alpha}}}
\def\veta{{\boldsymbol{\eta}}}

\def\vx{{\boldsymbol{x}}}
\def\vy{{\boldsymbol{y}}}
\def\vz{{\boldsymbol{z}}}

\def\rx{{\textnormal{x}}}
\def\ry{{\textnormal{y}}}
\def\rz{{\textnormal{z}}}
\def\ru{{\textnormal{u}}}

\def\1{\bm{1}}

\abstract{Tackling new machine learning problems with neural networks always means optimizing numerous hyperparameters that define their structure and strongly impact their performances. In this work, we study the use of goal-oriented sensitivity analysis, based on the Hilbert-Schmidt Independence Criterion (HSIC), for hyperparameter analysis and optimization. Hyperparameters live in spaces that are often complex and awkward. They can be of different natures (categorical, discrete, boolean, continuous), interact, and have inter-dependencies. All this makes it non-trivial to perform classical sensitivity analysis. We alleviate these difficulties to obtain a robust analysis index that is able to quantify hyperparameters’ relative impact on a neural network’s final error. This valuable tool allows us to better understand hyperparameters and to make hyperparameter optimization more interpretable. We illustrate the benefits of this knowledge in the context of hyperparameter optimization and derive an HSIC-based optimization algorithm that we apply on MNIST and Cifar, classical machine learning data sets, but also on the approximation of Runge function and Bateman equations solution, of interest for scientific machine learning. This method yields neural networks that are both  competitive and cost-effective.}

\keywords{Scientific machine learning, sensitivity analysis, Hilbert-Schmidt Independance Criterion, hyperparameter optimization, interpretability}

\maketitle

\section{Introduction}

Hyperparameter optimization is ubiquitous in machine learning, and especially in deep learning, where neural networks are often cluttered with many hyperparameters. For applications to real-world machine learning tasks, finding good hyperparameters carries high stakes. Indeed, hyperparameters have a strong impact on the prediction error of neural networks as well as on their cost efficiency, which is an important criterion in scientific computing. 

However, the balance between prediction accuracy and cost efficiency is difficult to find. For instance, recent breakthroughs owed to deep learning can be explained, among other reasons, by the ability to construct wider and deeper networks, while the computational cost of one neural network's prediction increases quadratically with its width and linearly with the depth \citep{deeplearningbook}. More generally, there are many hyperparameters in deep learning, all impacting the error and some of them impacting the execution time. It justifies carefully selecting them.

This selection can be fastidious for different reasons. i) The high number of hyperparameters by itself makes this problem challenging. ii) Their impact on error changes very often depending on the problem, so it is difficult to adopt general best practices and permanently recommend hyperparameter values for every machine learning problem. iii) Hyperparameters can be of very different natures, and have non-trivial relations, like conditionality or interactions.

In this paper, we tackle these problems by proposing a goal-oriented sensitivity analysis that we adapt and apply to complex hyperparameter search space. To this end, we select a powerful metric used for global sensitivity analysis, called Hilbert-Schmidt Independence Criterion (HSIC) \citep{hsicgretton}, which is a distribution dependence measure initially used for two-sample test problem \citep{two-sample-problem}. Once adapted to hyperparameter search space, HSIC gives insights into hyperparameters' relative importance for a given deep learning problem. It allows identifying which hyperparameters really matter in this situation, thereby addressing challenges i) and ii). Nonetheless, because of challenge iii), using HSIC in hyperparameter spaces is non-trivial. First, hyperparameters can be discrete (width of the neural network), continuous (learning rate), categorical (activation function), or boolean (batch normalization \citep{batchnorm}). Second, some hyperparameter's presence is conditional to others (moments decay rates specific to Adam optimizer \citep{adam}). Third, they can strongly interact (as shown in \cite{efficientnet}: in some cases, it is better to increase depth and width by a similar factor). The metric should be able to compare hyperparameters reliably in such situations. We introduce solutions to overcome this last challenge and illustrate them with some simple examples and with hyperparameter optimization for the approximation of Runge function. 

Once adapted to such complex spaces, we show that HSIC allows to understand hyperparameter relative importance better and to focus research efforts on specific hyperparameters. We also identify hyperparameters that have an impact on execution speed but not on the error.  Then, we introduce ways of reducing the hyperparameter's range of possible values to improve the stability of the training and neural network's cost efficiency. Finally, we propose an HSIC-based optimization methodology in two steps, one focused on essential hyperparameters and the other on remaining hyperparameters. It yields competitive and cost-effective neural networks for real-world machine learning problems: MNIST, Cifar10, and the approximation of the resolution of Bateman equations. This last problem is a physical data set of interest for scientific machine learning. 

\section{Sensitivity analysis as a new approach to hyperparameter optimization}

In this section, we describe the challenges of hyperparameter optimization. We emphasize the limits of classical hyperparameter optimization algorithms used to tackle these challenges and legitimate the approach of sensitivity analysis.

\subsection{Challenges of hyperparameter optimization}
\label{sect:pb-form}

\begin{figure}[!ht]
    \centering
    \includegraphics[width=0.8\linewidth]{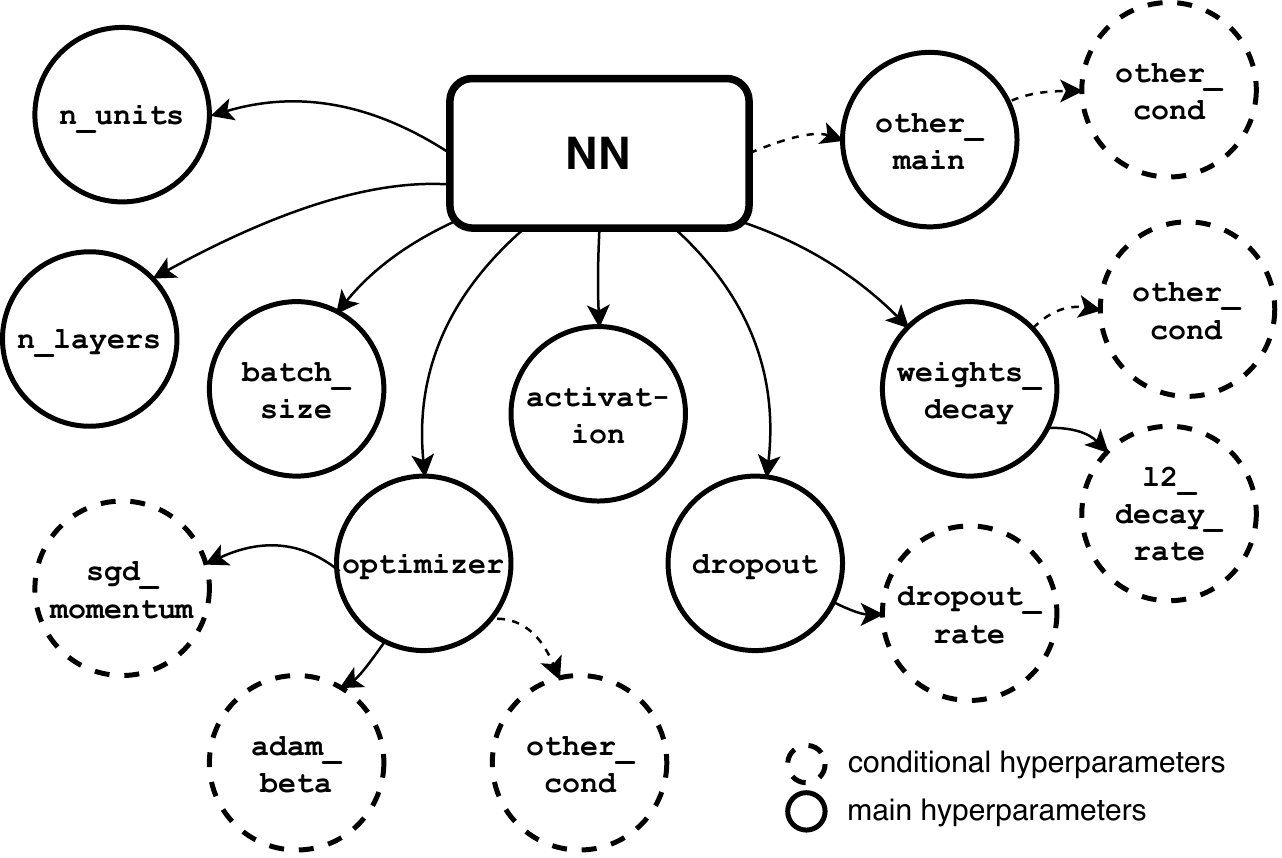}
    \caption{Example of hyperparameter space. 
    }
    \label{fig:tree}
\end{figure}

Let a neural network be described by $n_h$ hyperparameters $\rx_1, ..., \rx_{n_h}$ with $\rx_i \in \mathcal{X}_i$, $i \in \{1,...,n_h\}$ and $\boldsymbol{\sigma} = (\rx_{1}, ..., \rx_{n_h})$. We denote $F(\boldsymbol{\sigma})$ the error of the neural network on a test data set once trained on a training data set. The aim of hyperparameter optimization is to find $\boldsymbol{\sigma}^* = \underset{\boldsymbol{\sigma}}{\operatorname{argmin}} F(\boldsymbol{\sigma})$. Even if its formulation is simple, neural networks hyperparameter optimization is a challenging task because of the great number of hyperparameters to optimize, the computational cost for evaluating $F(\boldsymbol{\sigma})$ and the complex structure of hyperparameter spaces. Figure \ref{fig:tree} gives a graphical representation of a possible hyperparameter space and illustrates its complexity. Specific aspects to point out are the following ones : 

\begin{itemize}
    \item Hyperparameters do not live in the same measured space. Some are continuous ($\texttt{weights\_decay} \in [10^{-6}, 10^{-1}]$), some are integers ($\texttt{n\_layers} \in \{8,...,64\}$), others are categorical ($\texttt{activation} \in \{\texttt{relu},..., \texttt{sigmoid}\}$), or boolean ($\texttt{dropout} \in \{True, False\}$).
    \item They could interact with each others. For instance \texttt{batch\_size} adds variance on the objective function optimized by \texttt{optimizer}.
    \item Some hyperparameters are not involved for every neural networks configurations, e.g. \texttt{dropout\_rate} is not used when \texttt{dropout = False} or \texttt{adam\_beta} is only involved when \texttt{optimizer = adam}. In this case, we denote them as "conditional", otherwise we call them "main" hyperparameters. 
\end{itemize}

\subsection{Classical hyperparameter optimization techniques}

\indent Many techniques have been introduced to tackle the problem of hyperparameter optimization. Grid search or random search \citep{random-search} uniformly explore the search space. The main difference between the two methods is that hyperparameters values are chosen on a uniform grid for a grid search. These values are deterministic, whereas, for a random search, hyperparameters values are randomly sampled from a uniform distribution in a Monte Carlo fashion. The main advantages of random search over grid search are that it allows for more efficient exploration of the hyperparameter search and that it is not constrained to a grid, so it does not suffer from the curse of dimensionality - which is a problem here since the hyperparameters can be pretty numerous. On the other hand, these two methods can be costly because they require training a neural network for each hyperparameter configuration, so exploring the search space can be computationally very expensive.
    
Some methods aim at reducing the cost of such searches. For instance, Successive Halving \citep{succhalv} and Hyperband \citep{hyperband} train neural networks in parallel, like in grid search or random search, and stop their training after a certain number of epochs. Then, they choose the best half of neural networks and carry on the training only for these neural networks, for the same number of epochs, and so on. This procedure allows testing more hyperparameters values for the same computational budget. 

On the contrary, other methods are designed to improve the search quality with less training instances. Bayesian optimization, first introduced in \cite{bayesian_optim}, is based on the approximation of the loss function by a surrogate model. After an initial uniform sampling of hyperparameter configurations, the surrogate model is trained on these points and used to maximize an acquisition function. This acquisition function, often chosen to be expected improvement or upper confident bound \citep{boreview}, is supposed to lead to hyperparameter configurations that will improve the error. Therefore, it focuses the computation on potentially better hyperparameters values instead of randomly exploring the hyperparameter space. The surrogate model can be a Gaussian process \citep{hsic:spearmint}, a kernel density estimator \citep{algoho} or even a neural network \citep{dnnho}.

 Model-based hyperparameter optimization is not easily and naturally applicable to conditional or categorical hyperparameters that often appear when optimizing a neural network architecture. Such categorical hyperparameter can be the type of convolution layer for a convolutional neural network, regular convolution or depth-wise convolution \citep{hsic:depthwise}; and a conditional hyperparameter could be the specific parameters of each different convolution type. Neural architecture search explicitly tackles this problem. It dates back to evolutionary and genetic algorithms \citep{hsic:evo-nas} and has been the subject of many recent works. For instance, \cite{hsic:nasbot} models the architecture as a graph, or \cite{hsic:reinf,hsic:mnasnet} use reinforcement learning to automatically construct representations of the search space. See \cite{hsic:nas-survey} for an exhaustive survey of this field. Nevertheless, their implementation can be tedious, often involving numerous hyperparameters themselves.

Classical hyperparameter optimization methods handle hyperparameter optimization quite successfully. However, they are end-to-end algorithms that return one single neural network. The user does not interact with the algorithm during its execution. This lack of interactivity has many automating advantages but can bring some drawbacks. First, these methods do not give any insight on the relative importance of hyperparameters, whereas it may be of interest in the first approach to a machine learning problem. They are black boxes and not interpretable. Second, one could have other goals than the accuracy of a neural network, like execution speed or memory consumption. Some works like \cite{hsic:mnasnet} introduce multi-objective hyperparameter optimization, but it requires additional tuning of the hyperparameter optimization algorithm itself. Finally, there may be flaws in the hyperparameter space, like a useless hyperparameter that could be dropped but is included in the search space and becomes a nuisance for the optimization. This aspect is all the more problematic since some popular algorithms, like gaussian process-based Bayesian optimization, suffer from the curse of dimensionality. We can sum up the drawbacks as lack of interpretability, difficulties in a multi-objective context, and unnecessary search space complexity.

\subsection{Benefits of Sensitivity Analysis}

In this work, we alleviate these concerns by mixing hyperparameter optimization with hyperparameter analysis. In other words, we construct an approach to hyperparameter optimization that relies on assessing hyperparameter's effects on the neural network's performances. 

One powerful tool to analyze the effect of some input variables on the variability of a quantity of interest is sensitivity analysis \citep{hsic:futureofsas}. Sensitivity analysis consists of studying the sensitivity of the output of a function to its inputs. We could define this function as $F$ and its inputs as $\boldsymbol{\sigma}$. Then, it would be possible to make hyperparameter optimization benefit from characteristics of sensitivity analysis. Indeed, sensitivity analysis allows specifically:
\begin{itemize}
    \item Analyzing the relative importance of input variables for explaining the output, which helps to answer the lack of interpretability problem. We could better understand hyperparameters' impact on the neural network error.
    \item Selecting practically convenient values for input variables with a limited negative impact on the output. It simplifies the multi-objective approach since we could, for instance, select values that improve execution speed with a limited impact on the neural network error.
    \item Identifying where to efficiently put research efforts to improve the output, which answers the unnecessary search space complexity problem. Indeed, we could focus on fewer hyperparameters to optimize by knowing which of them most impact the neural network error. 
\end{itemize}

\subsection{Goal-oriented sensitivity analysis}

Several types of sensitivity measures can be estimated after an initial sampling of input vectors and their corresponding output values.
The first type of metric gives information about the contribution of an input variable to the output based on variance analysis. The most common metric used for that purpose are Sobol indices \citep{Sobol}, but they only assess the contribution of variables to the output variance. Goal-oriented Sobol indices \citep{contrast} or uncertainty importance measure \citep{borgonovo} construct quantities based on the output whose variance analysis gives more detailed information. However, computing these indices can be very costly since estimating them with an error of $\mathcal{O}(\frac{1}{\sqrt{n_s}})$ requires $(n_h+2)\times n_s$ sample evaluations \citep{hsic:saltelli-doe}, which can be prohibitive for hyperparameter analysis. 
Another type of metrics, called dependence measures, assesses the dependence between $\rx_i$ (that can be a random vector) and the output $F(\boldsymbol{\sigma})$  \citep{gsahsic}. It relies on the claim that the more $\rx_i$ is independent of $F(\boldsymbol{\sigma})$, the less important it is to explain it. Dependence measures are based on dissimilarity measures between $\mathbb{P}_{\rx_i}\mathbb{P}_{\ry}$ and $\mathbb{P}_{\rx_i,\ry}$, where $\rx_i \sim \mathbb{P}_{\rx_i}$ and $\ry = F(\boldsymbol{\sigma}) \sim \mathbb{P}_{\ry}$, since  $ \mathbb{P}_{\rx_i \ry} = \mathbb{P}_{\rx_i}\mathbb{P}_{\ry}$ when $\rx_i$ and $\ry$ are independent. In \cite{gsahsic}, the author gives several examples of indices based on dissimilarity measures like $f$-divergences \citep{fdivergence} or integral probability metrics \citep{IPM}. These indices are easier and less expensive to estimate ($n_s$ training instances instead of $(n_h+2)\times n_s$) than variance-based measures since they only need a simple Monte Carlo design of experiment. 

This work focuses on a specific dependence measure, known as the Hilbert-Schmidt Independence Criterion (HSIC). The following sections are dedicated to the description of HSIC and its adaptation to hyperparameter optimization.

\section{HSIC-based goal-oriented sensitivity analysis}

In this section, we recall the definition of Hilbert Schmidt Independence Criterion, how to use it in practice, and how to adapt it in order to perform goal-oriented sensitivity analysis.

\subsection{From Integral Probability Metrics to Maximum Mean Discrepancy}\label{mmd}

Let $\rvx$ and $\rvy$ be two random variables of probability distribution $\mathbb{P}_{\rvx}$ and $\mathbb{P}_{\rvy}$ defined in $\cal{X}$. \cite{two-sample-problem} show that distributions $\mathbb{P}_{\rvx} = \mathbb{P}_{\rvy}$ if and only if $\mathbb{E}[f(\rvx)] - \mathbb{E}[f(\rvy)] = 0$ for all $f \in C(\cal{X})$, where $C(\cal{X})$ is the space of bounded continuous functions on $\cal{X}$.
This lemma explains the intuition behind the construction of Integral Probability Metrics (IPM) \citep{IPM}. \\

Let $\cal{F}$ be a class of functions, $f:\cal{X} \rightarrow \mathbb{R}$. An IPM $\gamma$ is defined as

\begin{equation}\label{mmdf}
    \gamma(\mathcal{F},\mathbb{P}_{\rvx}, \mathbb{P}_{\rvy}) = \underset{f \in \mathcal{F}}{\operatorname{sup}} \vert \mathbb{E}[f(\rvx)] - \mathbb{E}[f(\rvy)]\vert .
\end{equation}

The Maximum Mean Discrepancy (MMD) can be defined as an IPM restricted to a class of functions $\mathcal{F}_{\mathcal{H}}$ defined on the unit ball of a Reproducing Kernel Hilbert Space (RKHS) $\mathcal{H}$ of kernel $k: \mathcal{X}^2 \rightarrow \mathbb{R} $. In \cite{hsicgretton}, this choice is motivated by the capacity of RKHS to embed probability distributions efficiently. The authors define $\mu_{\rvx}$ such that $\mathbb{E}(f(\rvx)) = \langle f, \mu_{\rvx}\rangle_\mathcal{H} $ as the mean embedding of $\mathbb{P}_{\rvx}$. Then, $\gamma_k^2(\mathbb{P}_{\rvx}, \mathbb{P}_{\rvy }) $ can be written

\begin{equation}
    \begin{split}
        \gamma_k^2 (\mathbb{P}_{\rvx}, \mathbb{P}_{\rvy }) &= \| \mu_{\rvx} - \mu_{\rvy } \|^2_{\mathcal{H}}.\\
         &= \int \int k(\vx, \vx) (p_{\rvx}(\vx) - p_{\rvy }(\vx))(p_{\rvx}(\vy) - p_{\rvy }(\vy))d\vx d\vx'\\
        &= \mathbb{E}[k(\rvx, \rvx')] + \mathbb{E}[k(\rvy, \rvy')] -2 \mathbb{E}[k(\rvx, \rvy)],
    \end{split}\label{gamma}
\end{equation}

\noindent where $p_{\rvx}(\vx)dx = d\mathbb{P}_{\rvx}$, $p_{\rvy }(\vy)dy = d\mathbb{P}_{\rvy }$,  $\rvx, \rvx'$ are iid (independent and identically distributed) and $\rvy, \rvy'$ are iid. After a Monte Carlo sampling of $\{\vx_1, ..., \vx_{n_s}\}$ and $\{\vy_1,...,\vy_{n_s}\}$, $\gamma_k^2 (\mathbb{P}_{\rvx}, \mathbb{P}_{\rvy })$ can thus be estimated by $ \hat{\gamma}_{k}^2(\mathbb{P}_{\rvx}, \mathbb{P}_{\rvy})$, with

\begin{equation}
        \hat{\gamma}_{k}^2(\mathbb{P}_{\rvx}, \mathbb{P}_{\rvy})=  \sum_{j=1}^{n_s} \sum_{l=1}^{n_s} k(\vx_{j}, \vx_{l}) 
        + \sum_{j=1}^{n_s} \sum_{l=1}^{n_s} k(\vy_{j}, \vy_{l}) 
        -2 \sum_{j=1}^{n_s} \sum_{l=1}^{n_s} k(\vx_{j}, \vy_{l}),
\end{equation}

\noindent and $\hat{\gamma}_{k}^2(\mathbb{P}_{\rvx}, \mathbb{P}_{\rvy})$ being an unbiased estimator, its standard error can be estimated using the empirical variance of $\hat{\gamma}_{k}^2(\mathbb{P}_{\rvx}, \mathbb{P}_{\rvy})$.

\subsection{The kernel choice}

\Eqref{gamma} involves to choose a kernel $k$. In practice, $k$ is chosen among a class of kernels that depends on a set of parameters $\mathbf{h} \in \mathbf{H}$, where $\mathbf{H}$ is a kernel parameter space. We therefore temporarily denote the kernel by $k_{\mathbf{h}}$. Examples of kernels are the Gaussian Radial Basis Function $k_h : (\vx,\vy) \rightarrow   \exp(-\frac{\|\vx - \vy\|^2}{2h^2})$ or the Matérn function  $k_\mathbf{h} : (\vx,\vy) \rightarrow   \sigma^2 \frac{2^{1-\nu}}{\Gamma(\nu)} \Big( \sqrt{2\nu}\frac{\|\vx - \vy\|}{\eta} \Big)^{\nu} K_{\nu}\Big( \sqrt{2\nu}\frac{\|\vx - \vy\|}{\eta} \Big)$, where $\mathbf{h}=\{\sigma, \nu, \eta\}$, $\Gamma$ is the gamma function and $K_{\nu}$ is the modified Bessel function of the second kind.   In \cite{MMD-generalized}, the authors study the choice of the kernel, and more importantly of the kernel parameters $\mathbf{h}$. They state that, for the comparison of probabilities $\mathbb{P}_{\rvx}$ and $\mathbb{P}_{\rvy}$,  the final parameter $\mathbf{h}^*$ should be chosen such that

\begin{equation}
    \gamma_{k_{\mathbf{h}^*}}^2(\mathbb{P}_{\rvx}, \mathbb{P}_{\rvy}) = \underset{\textbf{h} \in \mathbf{H}}{\sup}\gamma_{k_{\mathbf{h}}}^2(\mathbb{P}_{\rvy}, \mathbb{P}_{\rvy}).
\end{equation}

The authors suggest focusing on unnormalized kernel families, like Gaussian Radial Basis Functions $ \big\{k_h : (\vx,\vy) \rightarrow  \exp(-\frac{\|\vx - \vy\|^2}{2h^2}), h \in (0, \infty) \big\}$, also used in \cite{gsahsic}, for which they demonstrate that $\hat{\gamma}_{k_{\mathbf{h}^*}}^2(\mathbb{P}_{\rvx}, \mathbb{P}_{\rvy})$, defined as 

\begin{equation}
        \hat{\gamma}_{k_{\mathbf{h}^*}}^2(\mathbb{P}_{\rvx}, \mathbb{P}_{\rvy})= \underset{\mathbf{h} \in \mathbf{H}}{\sup}\Bigg[ \sum_{j=1}^{n_s} \sum_{l=1}^{n_s} k_{\mathbf{h}}(\vx_{j}, \vx_{l}) 
        + \sum_{j=1}^{n_s} \sum_{l=1}^{n_s} k_{\mathbf{h}}(\vy_{j}, \vy_{l}) 
        -2 \sum_{j=1}^{n_s} \sum_{l=1}^{n_s} k_{\mathbf{h}}(\vx_{j}, \vy_{l})\Bigg],
\end{equation}

\noindent is a consistent estimator of $\gamma_{k_{\mathbf{h}^*}}^2(\mathbb{P}_{\rvx}, \mathbb{P}_{\rvy})$. It is thus possible to choose $\mathbf{h}$ by maximizing $\hat{\gamma}_{k_{\mathbf{h}}}^2(\mathbb{P}_{\rvx}, \mathbb{P}_{\rvy})$ with respect to $\mathbf{h}$. Therefore, in this work, we use Gaussian Radial Basis Functions kernel. Once $\mathbf{h}^*$ is chosen, the approximation error of $\hat{\gamma}_{k_{\mathbf{h}^*}}^2(\mathbb{P}_{\rvx}, \mathbb{P}_{\rvy})$ can also be estimated like in Section \ref{mmd}. It is important to note that $\gamma_{k_{\mathbf{h}}}^2(\mathbb{P}_{\rvx}, \mathbb{P}_{\rvy})$ can be estimated in a $\mathcal{O}(n_s^2)$ computational complexity, which is not expensive given usual values of $n_s$ in hyperparameter optimization context. The total complexity of the minimization process depends on the minimization algorithm, but since the optimization problem is low dimensional - there is never more than a handful of kernel parameters - the whole process is always cost effective. To simplify the notations, we denote $k_{\mathbf{h}^*}$ by $k$ in the following sections.

\subsection{Hilbert-Schmidt Independence Criterion (HSIC) for goal-oriented sensitivity analysis}

Let $\rvx \in \mathcal{X}$ and $\rvy \in \mathcal{Y}$, and $\mathcal{G}$ the RKHS of kernel $k : \mathcal{X}^2 \times \mathcal{Y}^2 \rightarrow \mathbb{R}$. HSIC can be written 

\begin{equation}\label{eq:hsicdef}
    HSIC(\rvx,\rvy) = \gamma_{k}^2 (\mathbb{P}_{\rvx \rvy}, \mathbb{P}_{\rvx}\mathbb{P}_{\rvy}) = \| \mu_{\rvx \rvy} - \mu_{\rvx}\mu_{\rvy} \|_{\mathcal{G}}.
\end{equation}

Then, HSIC measures the distance between $\mathbb{P}_{\rvx \rvy}$ and $\mathbb{P}_{\rvy}\mathbb{P}_{\rvx}$ embedded in $\mathcal{H}$. Indeed, since $\rvx \perp \rvy \Rightarrow \mathbb{P}_{\rvx \rvy} = \mathbb{P}_{\rvy}\mathbb{P}_{\rvx}$, the closer these distributions are, in the sense of $\gamma_{k}$, the more independent they are.

In \cite{hsic}, the authors present a goal-oriented sensitivity analysis by focusing on the sensitivity of $F$ w.r.t. $\rvx_i$ when $\ry = F(\rvx_1, ...,\rvx_{n_h}) \in \mathbf{Y}$, with $\mathbf{Y} \subset \mathbb{R}$. The sub-space $\mathbf{Y}$ is chosen based on the goal of the analysis. In the context of optimization, for instance, $\mathbf{Y}$ is typically chosen to be the best percentile of $Y$. To achieve this, the authors introduce a new random variable, $\rz = \1_{\ry \in \mathbf{Y}}$. Then,

\begin{equation}\label{eq:hsicz}
    HSIC(\rvx_i,\rz) = \mathbb{P}(\rz = 1)^2 \times \gamma^2_k(\mathbb{P}_{\rvx_i\vert\rz=1}, \mathbb{P}_{\rvx_i}),
\end{equation}

\noindent so $HSIC(\rvx_i, \rz)$ measures the distance between $\rvx_i$ and $\rvx_i\vert\rz=1$ (to be read $\rvx_i$ conditioned to $\rz=1$) and can be used to measure the importance of $\rvx_i$ to reach the sub-space $\mathbf{Y}$ with $F$. Using the expression of $\gamma_k$ given by \eqref{gamma}, its exact expression is

\begin{equation}\label{eq:hsic}
    HSIC(\rvx_i,\rz) = \mathbb{P}(\rz = 1)^2 \Big[\mathbb{E}[k(\rvx_i, \rvx'_i)] + \mathbb{E}[k(\rvz, \rvz')] -2 \mathbb{E}[k(\rvx_i, \rvz)] \Big],
\end{equation}

\noindent where $\rvx_i, \rvx'_i$ are iid and $\rvz, \rvz'$ are iid. It is estimated for each $\rvx_i$ using Monte Carlo estimators denoted by $S_{ \rvx_i , \mathbf{Y}}$, based on samples $\{\vx_{i,1},...,\vx_{i,n_s}\}$ from $\rvx_i \sim d\mathbb{P}_{\rvx_i}$ and corresponding $\{z_1,...,z_{n_s}\}$ drawn from $\rz$. The estimator $S_{ \rvx_i , \mathbf{Y}}$ is defined as

\begin{equation}\label{eq:sxy}
    \begin{split}
        S_{\rvx_i,\mathbf{Y}}= \mathbb{P}(\rz = 1)^2 \Bigg[  \frac{1}{m^2} & \sum_{j=1}^{n_s} \sum_{l=1}^{n_s} k(\vx_{i,j}, \vx_{i,l})\delta(z_{j} =1)\delta(z_{l} =1) \\
        + \frac{1}{{n_s}^2} &\sum_{j=1}^{n_s} \sum_{l=1}^{n_s} k(\vx_{i,j}, \vx_{i,l}) \\
        - \frac{2}{n_s m}&\sum_{j=1}^{n_s} \sum_{l=1}^{n_s} k(\vx_{i,j}, \vx_{i,l}) \delta(z_{l} =1)\Bigg],
    \end{split}
\end{equation}

\noindent with $m=\sum_k^{{n_s}} \delta(z_{k} =1) $ and $\delta(a) = 1$ if $a$ is True and $0$ otherwise. We use this metric in the following. This section mainly summed up the mathematics on which the sensitivity indices are based and how they are used in practice in a sensitivity analysis context. The following section is devoted to the application of HSIC in hyperparameter spaces.\\

\section{Application of HSIC to hyperparameter spaces}\label{sect:hsicforha}

HSIC has two advantages that make it stand out from other sensitivity indices and make it particularly suitable to hyperparameter spaces. First, \eqref{eq:sxy} emphasizes that it is possible to estimate HSIC using simple Monte Carlo estimation. Hence, in the context of hyperparameter optimization, such indices can be estimated after a classical random search.  
Secondly, using \eqref{eq:hsicz}, HSIC allows to perform goal-oriented sensitivity analysis easily, i.e. to assess the importance of each hyperparameter for the error to reach a given $\mathbf{Y}$. For hyperparameter analysis, $\mathbf{Y}$ can be chosen to be the sub-space for which $F(\rx_1,...,\rx_{n_h})$ is in the best percentile $p$ of a metric ($L_2$ error, accuracy,...), say $p=10\%$. Then, the quantity $S_{ \rx_i , \mathbf{Y}}$ measures the importance of each hyperparameter $\rx_i$ for obtaining the $10\%$ best neural networks.

However, one cannot use HSIC as is in hyperparameter analysis. Indeed, hyperparameters do not live in the same measured space, they could interact with each other, and some are not directly involved for each configuration. In the following sections, we suggest some original solutions to these issues. 

To illustrate the performances of these solutions, we consider a toy example which is the approximation of Runge function $r:x \rightarrow \frac{1}{1 + 15x^2}$, $x\in [-1, 1]$ by a fully connected neural network. This approximation problem is a historical benchmark of approximation theory. We consider $n_h=14$ different hyperparameters (see \textbf{\hyperref[appA]{Appendix A}} for details). We randomly draw $n_s=10000$ hyperparameter configurations and perform the corresponding training on $11$ training points. We record the test error on a test set of $1000$ points. All samples are equally spaced between $0$ and $1$. 

In Section \ref{normalization}, we introduce a transformation to deal with hyperparameters that do not live in the same measured space. Then, in Section \ref{sect:interactions} we explain how to use HSIC to evaluate hyperparameters' interactions. Finally, in Section \ref{sect:cond} we deal with conditionality between hyperparameters. 

\subsection{Normalization of hyperparameter space}\label{normalization}

Hyperparameters can be defined in very different spaces. For instance, the activation function is a categorical variable that can be \texttt{relu}, \texttt{sigmoid} or \texttt{tanh}, \texttt{dropout\_rate} is a continuous variable between $0$ and $1$ while \texttt{batch\_size} is an integer that can go from 1 to hundreds. Moreover, it may be useful to sample hyperparameters with a non-uniform distribution (e.g. log-uniform for \texttt{learning\_rate}). Doing so affects HSIC value and its interpretation, which is undesirable since this distribution choice is arbitrary and only relies on practical considerations. Let us illustrate this phenomenon in the following example.\\

\begin{example}
\label{ex1} Let $f: [0,2]^2 \rightarrow \{0,1\}$ such that 

\begin{equation*}
       f(\rx_1, \rx_2) = 
   \begin{dcases}
    1 & \text{if  } \rx_1 \in [0,1], \rx_2 \in [0,1],\\
    0 & \text{otherwise.  } 
    \end{dcases}
\end{equation*}
\end{example}
Suppose we want to assess the importance of $\rx_1$ and $\rx_2$ for reaching the goal $f(\rx_1, \rx_2)=1$ without knowing $f$. In the formalism of the previous section, we have $\mathbf{Y} = \{1\}$. Regarding its definition, $\rx_1$ and $\rx_2$ are equally important for $f$ to reach $\mathbf{Y}$, due to their symmetrical effect. Let $\rx_1 \sim \mathcal{N}(1, 0.1, [0, 2])$ (normal distribution of mean $1$ and variance $0.1$ truncated between $0$ and $2$) and $\rx_2 \sim \mathcal{U}[0,2]$. We compute $S_{ \rx_1 , \mathbf{Y}}$ and $S_{ \rx_2 , \mathbf{Y}}$ with $n_s=10000$ points and display their value in Table \ref{tab:norm1}. The values of $S_{ \rx_1 , \mathbf{Y}}$ and $S_{ \rx_2 , \mathbf{Y}}$ are quite different. It is natural since their chosen initial distribution is different. However, this choice has nothing to do with their actual importance; it should not have any impact on the importance measure. Here, we could erroneously conclude that $\rx_2$ is more important than $\rx_1$.

\begin{table}[!ht]
  \centering
\begin{minipage}{0.49\linewidth}
        \raggedright
        \begin{tabular}{lll}
             & $\rx_1$ & $\rx_2$   \\
            \hline
            $S_{\rx, \mathbf{Y}} (\times 10^{-2})$  & $1.17 \pm 0.05$ & $1.55  \pm 0.05$    \\
            
        \end{tabular}
        \caption{\label{tab:norm1} $S_{\rx, \mathbf{Y}} $ values for $\rx_1$ and $\rx_2$}
\end{minipage}
\begin{minipage}{0.49\linewidth}
        \raggedleft
        \begin{tabular}{lll}
             &  $\ru_1$ & $\ru_2$   \\
            \hline
            $S_{\rx, \mathbf{Y}} (\times 10^{-2})$  & $1.54 \pm 0.05$ & $1.55  \pm 0.05$   \\
            
        \end{tabular}
        \caption{\label{tab:norm2} $S_{\rx, \mathbf{Y}} $ values for $\ru_1$ and $\ru_2$}
\end{minipage}
\end{table}

This example shows that we have to make $S_{ \rx_i , \mathbf{Y}}$ and $S_{ \rx_j, \mathbf{Y}}$ comparable in order to say that hyperparameter $\rx_i$ is more important than hyperparameter $\rx_j$. Indeed, if $\rx_i$ and $\rx_j$ do not follow the same distribution or $\mathcal{X}_i \neq \mathcal{X}_j$, it may be irrelevant to compare them directly. We need a method to obtain values for $S_{ \rx_i , \mathbf{Y}}$ that are robust to the choice of $d\mathbb{P}_{\rx_i}$. To tackle this problem, we introduce the following approach for comparing variables with HSIC. Let $\Phi_i$ be the CDF of $\rx_i$. We have that $\Phi_i(\rx_i) =\ru_i$, with $\ru_i \sim \mathcal{U}[0,1]$. After an initial Monte Carlo sampling of hyperparameter $\rx_i$, which can be a random search, we can apply $\Phi_i$ to each input point to obtain $\ru_i$ corresponding to $\rx_i$ with $\ru_i$ iid, so living in the same measured space. Yet, one must be aware that to obtain $\ru_i \sim \mathcal{U}[0,1]$, its application is different for continuous and discrete variables: 

\begin{itemize}
    \item for continuous variables, $\Phi_i(\rx_i)$ is a bijection between $\mathcal{X}_i$ and $[0,1]$ so $\Phi_i$ can be applied on draws from $\rx_i$.
    \item For categorical, integer or boolean variables,  $\Phi_i(\rx_i)$ is not a bijection between $\mathcal{X}_i$ and $[0,1]$. Suppose that $\rx_i$ is a discrete variable with $p$ possible values $\{\rx_{i}[1],...,\rx_{i}[p]\}$, each with probability $w_p$. Let us encode $\{\rx_{i}[1],...,\rx_{i}[p]\}$ by $\{1,...,p\}$. Then, $\Phi_i(\rx_i) = \sum_{j=1}^p w_j \1_{[ \rx_i \leq j]}(\rx_i)$. When $\Phi_i$ is applied as is, $\Phi_i(\rx_i)$ is not uniform. To overcome that, one can simply use $\ru_i = \sum_{j=1}^p \mathcal{U}[\sum_{k < j}w_k, \sum_{k < j+1}w_k]\delta(\rx_i=j)$. This trick, introduced in \cite{categ}, is commonly used in Monte Carlo resolution of Partial Differential Equations \citep{MCcdf}. As a result, $\ru_i \sim \mathcal{U}[0,1]$.
\end{itemize}

Finally, all we have to do is to sample $\rx_i$, like in a classical random search, following the distribution we want, and then apply $\Phi_i$ to obtain $\ru_i$. The corresponding HSIC estimation is $S_{ \ru_i, \mathbf{Y}}$. It only involves $\ru_i$ and $\ru_i\vert\rz=1$ and since $\ru_i$ are iid, the comparison of different $S_{ \ru_i, \mathbf{Y}}$ becomes relevant. Coming back to the previous example, Table \ref{tab:norm2} displays values of $S_{ \ru_1, \mathbf{Y}}$ and $S_{ \ru_2, \mathbf{Y}}$. This time, the value is the same, leading to the correct conclusion that both variables are equally important. Note that in the following, we denote $S_{ \ru_i, \mathbf{Y}}$ by $S_{ \rx_i , \mathbf{Y}}$ for clarity but always resort to this transformation.

\begin{figure}[!ht]
  \centering
  \includegraphics[width=0.6\linewidth]{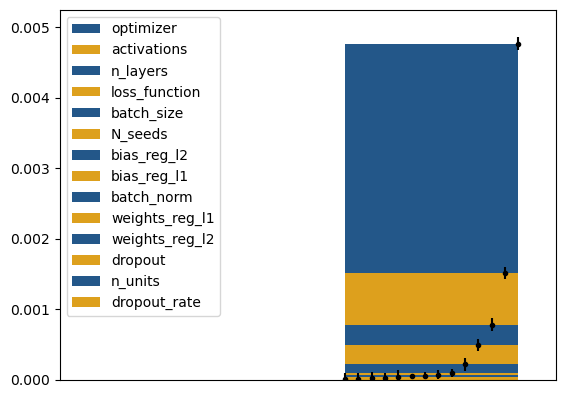}
  \caption{Comparison of $S_{ \rx_i , \mathbf{Y}}$ for hyperparameters in Runge approximation problem. The hyperparameters are ordered from the most important (top of the legend) to the least important (bottom of the legend), and their value is graphically represented in a stacked bar plot following the same order.}
  \label{fig:hsicrunge}
\end{figure}

Let us apply this methodology to the Runge approximation hyperparameter analysis problem. Figure \ref{fig:hsicrunge} displays a comparison between $S_{ \rx_i , \mathbf{Y}}$ for hyperparameters of the Runge approximation problem, with $\mathbf{Y}$ the set of the $10\%$ best neural networks. For readability, we order $\rx_i$ by $S_{ \rx_i , \mathbf{Y}}$ value in the legend and the figure. We also display black error bars corresponding to HSIC estimation standard error.  This graphic highlights that \texttt{optimizer} is by far the most important hyperparameter for this problem, followed by \texttt{activation}, \texttt{loss\_function} and \texttt{n\_layers}. Other hyperparameters may be considered non-impactful because their $S_{ \rx_i , \mathbf{Y}}$ values are low. Besides, these values are lower than the error evaluation. It could be only noise, and therefore these hyperparameters can not be ordered on this basis.

\subsection{Interactions between hyperparameters}\label{sect:interactions}

If $S_{ \rx_i , \mathbf{Y}}$ is low, it means that $\mathbb{P}_{\rx_i}$ and $\mathbb{P}_{\rz}$ are close to independent. We could want to conclude that $\rx_i$ has a limited impact on $Y$. However, $\rx_i$ may have an impact due to its interactions with some of the other hyperparameters. In other words, let $\rx_i$ and $\rx_j$ be two variables, it can happen that $S_{ \rx_i , \mathbf{Y}}$ and $S_{\rx_j,\mathbf{Y}}$ are low while $S_{(\rx_i, \rx_j),\mathbf{Y}}$ is high. \\

\begin{figure}[!ht]
   \begin{subfigure}{0.245\textwidth}
     \includegraphics[width=1.0\linewidth]{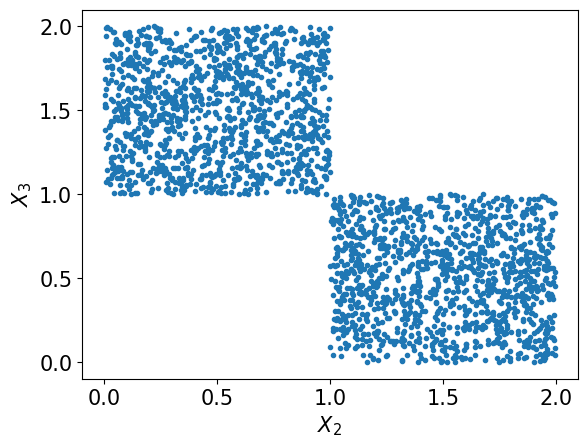}
   \end{subfigure}
   \begin{subfigure}{0.245\textwidth}
     \includegraphics[width=1.0\linewidth]{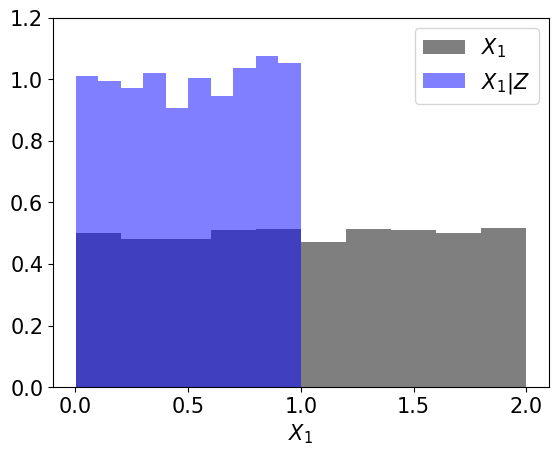}
   \end{subfigure}
      \begin{subfigure}{0.245\textwidth}
     \includegraphics[width=1.0\linewidth]{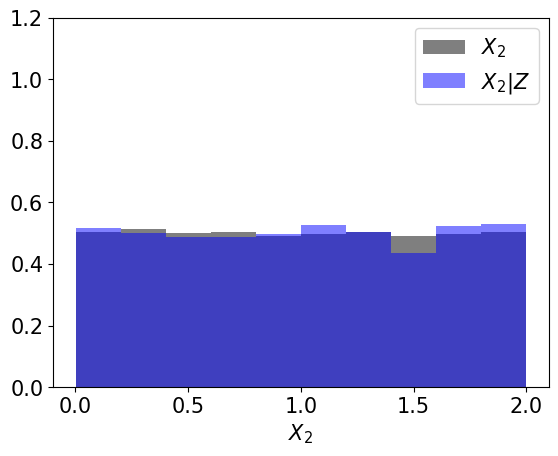}
   \end{subfigure}
      \begin{subfigure}{0.245\textwidth}
     \includegraphics[width=1.0\linewidth]{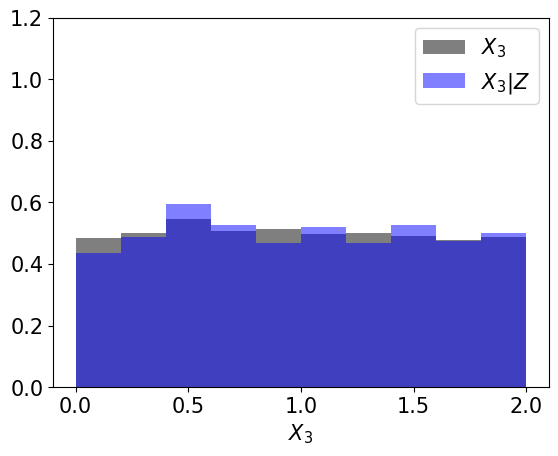}
   \end{subfigure}
   \caption{\textbf{From left to right:} \textbf{1} - Pairs of $(\rx_2\vert\rz=1, \rx_3\vert\rz=1)$. \textbf{2} - Histogram of $\rx_1$ and  $\rx_1\vert\rz=1$. \textbf{3} - Histogram of $\rx_2$ and  $\rx_2\vert\rz=1$. \textbf{4} - Histogram of $\rx_3$ and  $\rx_3\vert\rz=1$.}
   \label{fig:flaws}
\end{figure}

\begin{example}\label{ex2}

For instance let $f: [0,2]^3 \rightarrow \{0,1\}$ such that 

\begin{equation*}
       f(\rx_1, \rx_2, \rx_3) = 
   \begin{dcases}
    1 & \text{if  } \rx_1 \in [0,1], \rx_2 \in [1,2], \rx_3 \in [0,1],\\
    1 & \text{if  } \rx_1 \in [0,1], \rx_2 \in [0,1], \rx_3 \in [1,2],\\
    0 & \text{otherwise.  } 
    \end{dcases}
\end{equation*}
\end{example}  
In that case, let $\mathbf{Y} = \{1\}$, $\forall x \in [0,2]$ we have $p_{\rx_2\vert\rz=1}(x) = p_{\rx_2}(x)$ and  $p_{\rx_3\vert\rz=1}(x) = p_{\rx_3}(x)$. Hence, according to \eqref{eq:hsic} we have $HSIC(\rx_2,\rz) = HSIC(\rx_3,\rz) = 0$. However, we have

\begin{equation*}
    \begin{split}
        HSIC(\rx_1,\rz) &= \mathbb{P}(\rz = 1)^2 \int_{[0,2]^2} k(x,x') \big[p_{\rx_1\vert\rz=1}(x) - p_{\rx_1}(x)\big]\\ 
        &\;\;\;\; \times\big[p_{\rx_1\vert\rz=1}(x') - p_{\rx_1}(x')\big]dxdx'\\
        &= \frac{1}{8}\Big[ \int_{[0,1]\times[0,1]} k(x, x')dxdx' + \int_{[1,2]\times[1,2]} k(x, x')dxdx'  \\
        &\;\;\;\;-2\int_{[0,1]\times[1,2]} k(x, x')dxdx' \Big],
    \end{split}
\end{equation*}

\noindent so for non-trivial choice of $k$, $HSIC(\rx_1,\rz) \neq 0$. One could deduce that $\rx_1$ is the only relevant variable for reaching $\mathbf{Y}$, but in practice it is necessary to chose $\rx_2$ and $\rx_3$ carefully as well. For instance, if $\rx_1 \in[0,1]$, $f(\rx_1, \rx_2, \rx_3) = 1$ if $\rx_2 \in [1,2]$ and  $\rx_3 \in [0,1]$ but $f(\rx_1, \rx_2, \rx_3) = 0$ if $\rx_2 \in [1,2]$ and $\rx_3 \in [1,2]$. This is illustrated in Figure \ref{fig:flaws}, which displays the histograms of $\rx_1$ and $\rx_1\vert\rz=1$, $\rx_2$ and $\rx_2\vert\rz=1$, $\rx_3$ and $\rx_3\vert\rz=1$, obtained from $10000$ points $(\rx_1, \rx_2, \rx_3)$ sampled uniformly in the definition domain of $f$.\\

Histograms are the same for $\rx_2$, $\rx_2\vert\rz=1$  and $\rx_3$, $\rx_3\vert\rz=1$ (uniform between $0$ and $2$), but different for $\rx_1$, $\rx_1\vert\rz=1$. Therefore, HSIC being a distance measure between $\rx_1$ and $\rx_1\vert\rz=1$, it becomes intuitive that it will be high for $\rx_1$ and close to zero for $\rx_2$ and $\rx_3$, even if $\rx_2$ and $\rx_3$ are important as well because of their interaction. To assess this intuition, we compute $S_{\rx_1, \mathbf{Y}} $ , $S_{\rx_2, \mathbf{Y}} $, $S_{\rx_3, \mathbf{Y}} $ and  $S_{(\rx_2, \rx_3), \mathbf{Y}} $ after simulating $f$ for $n_s=2000$ points. We also compute  $S_{(\rx_4, \rx_5), \mathbf{Y}} $, with $\rx_4$ and $\rx_5$ two dummy variables, uniformly distributed, to have a reference for $S_{(\rx_2, \rx_3), \mathbf{Y}} $. The results can be found in Table \ref{tab:inter}. They show that $S_{\rx_1, \mathbf{Y}} $  and  $S_{(\rx_2, \rx_3), \mathbf{Y}} $ are of the same order while $S_{\rx_2, \mathbf{Y}} $, $S_{\rx_3, \mathbf{Y}} $ and $S_{(\rx_4, \rx_5), \mathbf{Y}} $ are two decades lower than $S_{\rx_1, \mathbf{Y}} $, which confirms that $S_{\rx, \mathbf{Y}} $ may be low while interactions are impactful.

\begin{table}[!ht]
\centering
    \begin{tabular}{llllll}
         & $\rx_1$ & $\rx_2$ & $\rx_3$ & $(\rx_2, \rx_3)$ & $(\rx_4, \rx_5)$  \\
        \hline
        $S_{\rx, \mathbf{Y}} $  & $1.51 \times 10^{-2}$ & $6.26 \times 10^{-6}$ & $1.64 \times 10^{-5}$ & $3.47 \times 10^{-3}$ &  $3.70 \times 10^{-6}$  \\
        
    \end{tabular}
    \caption{\label{tab:inter} $S_{\rx, \mathbf{Y}} $ values for variables of the experiment}
\end{table}

\begin{figure}
  \centering
  \includegraphics[width=0.7\linewidth]{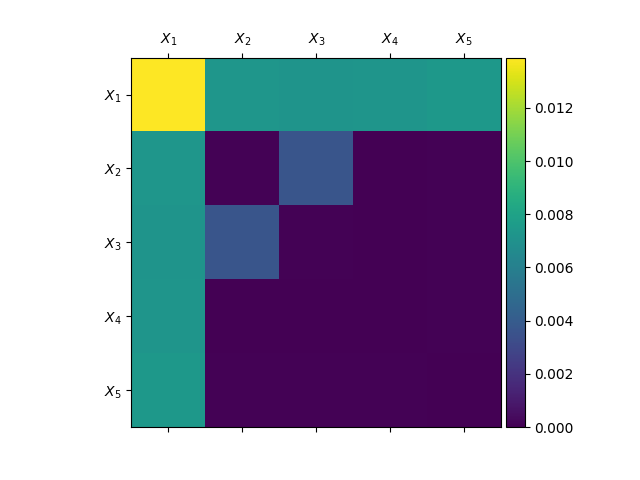}
  \caption{$S_{\rx, \mathbf{Y}} $ for each pair of variable.}
  \label{fig:interactions}

\end{figure}

 Additionally, we display the $S_{(\rx_i, \rx_j), \mathbf{Y}} $ for each pair of variable $\rx_i$ and $\rx_j$ on Figure \ref{fig:interactions}. We can see that for variables other than $\rx_1$, $S_{(\rx_i, \rx_j), \mathbf{Y}} $ is high only for $i=2$ and $j=3$. This example shows that it is necessary to compute $S_{\rx, \mathbf{Y}} $ of joint variables to perceive the importance of interactions between variables. 
 
\begin{figure}[!ht]
      \begin{subfigure}{0.5\textwidth}
     \includegraphics[width=1.0\linewidth]{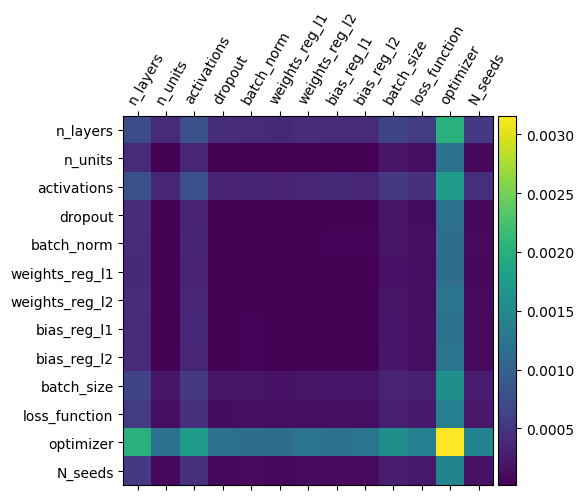}
     
     \caption{}
     \label{fig:inter_runge_1}
   \end{subfigure}
      \begin{subfigure}{0.5\textwidth}
     \includegraphics[width=1.0\linewidth]{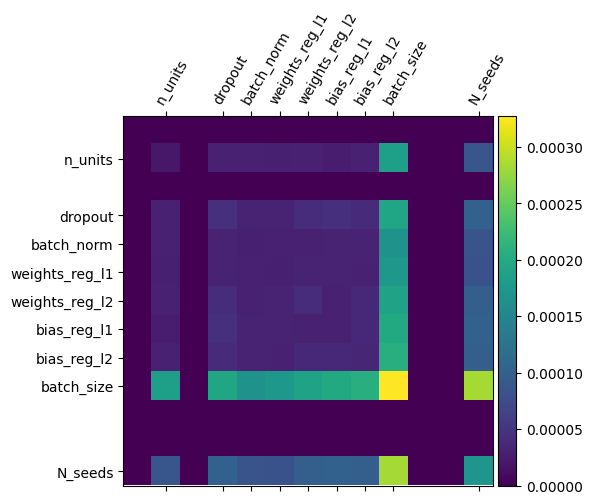}
     
      \caption{}
     \label{fig:inter_runge_2}
   \end{subfigure}
   \caption{(a) $S_{(\rx_i, \rx_j), \mathbf{Y}} $ for each pair of hyperparameters. (b) $S_{(\rx_i, \rx_j), \mathbf{Y}} $ for each pair of hyperparameters, except for \texttt{optimizer}, \texttt{activation}, \texttt{n\_layers} and \texttt{loss\_function}. The grid can be read symmetrically with respect to the diagonal.}
\end{figure}

 The values are easy to interpret in this example because we know the behavior of the underlying function $f$. In practice, $S_{\rx_1, \mathbf{Y}} $ and $S_{(\rx_2,\rx_3), \mathbf{Y}} $ can not be compared because $(\rx_2,\rx_3)$ and $\rx_1$ do not live in the same measured space ($\mathcal{X}_2 \times \mathcal{X}_3$ and $\mathcal{X}_1$ respectively). Moreover, like we see on Figure \ref{fig:interactions}, $S_{(\rx_i, \rx_j), \mathbf{Y}} $ is always the highest when $i=1$, regardless of $j$. In fact, if for a given variable $\rx_i$, $S_{\rx_i, \mathbf{Y}} $ is high, so will be $S_{(\rx_i,\rx_j), \mathbf{Y}} $ for any other variable $\rx_j$. Hence, care must be taken to only compare interactions of low $S_{\rx, \mathbf{Y}} $ variables with each others, and not with high $S_{\rx, \mathbf{Y}} $ variables. Coming back to Runge approximation example, Figure \ref{fig:inter_runge_1} displays the  $S_{(\rx_i, \rx_j), \mathbf{Y}} $ for each pair of hyperparameters, and Figure \ref{fig:inter_runge_2} for each pair of hyperparameters, except for the impactful hyperparameters \texttt{optimizer}, \texttt{activation}, \texttt{n\_layers} and \texttt{loss\_function}.

Figures \ref{fig:inter_runge_1} and \ref{fig:inter_runge_2} illustrate the remarks of the previous section. First, if we only look for interactions on Figure \ref{fig:inter_runge_1}, we would conclude that the most impactful hyperparameters are the only one to interact, and that they only interact with each others. Figure \ref{fig:inter_runge_2} shows that this conclusion is not true. Hyperparameter \texttt{batch\_size} is the $5$-th most impactful hyperparameter, and like we can see in Figure \ref{fig:hsicrunge}, is slightly above the remaining hyperparameters. It is normal that $S_{( \texttt{batch\_size}, \rx_j), \mathbf{Y}} $ is high, with $\rx_j$ every other hyperparameters. However, $S_{( \texttt{batch\_size}, \texttt{n\_units}), \mathbf{Y}} $ is higher, whereas \texttt{n\_units} is the $13$-th most impactful hyperparameter. This means that \texttt{batch\_size} interacts with \texttt{n\_units} in this problem, i.e. that when considered together, they contribute to explain the best results.

\subsection{Conditionality between hyperparameters}\label{sect:cond}

Conditionality between hyperparameters, which often arises in Deep Learning, is a non-trivial challenge in hyperparameter optimization. For instance, hyperparameter "$\texttt{dropout\_rate}$" will only be involved when hyperparameter "$\texttt{dropout}$" is set to $\texttt{True}$. Classically, two approaches can be considered. The first \textbf{(i)} splits the hyperparameter optimization between disjoint groups of hyperparameters that are always involved together, like in \cite{algoho}. Then, two separate instances of hyperparameter optimization are created, one for the main hyperparameters and another for $\texttt{dropout\_rate}$. The second \textbf{(ii)} considers these hyperparameters as if they were always involved, even if they are not, like in \cite{hsic:bohb}. In that case, $\texttt{dropout\_rate}$ is always assigned a value even when $\texttt{dropout} = \texttt{False}$, and these dummy values are used in the optimization. First, we explain why these two approaches are not suited to our case. Then we propose a third approach \textbf{(iii)}.\\

\textbf{\textbf{(i)}}  The first formulation splits the hyperparameters between disjoints sets of hyperparameters whose value and presence are involved jointly in a training instance. In Runge approximation hyperparameter analysis, since \texttt{dropout\_rate} is the only conditional hyperparameter, it would mean to split the hyperparameters between two groups: $\{\texttt{dropout\_rate}\}$ and another containing all the others. This splitting approach is not suited to HSIC computation because it produces disjoints sets of hyperparameters, while we would want to measure the importance of every hyperparameter and compare it to each other hyperparameter. Here, \texttt{dropout\_rate} could not be compared to any other hyperparameters.\\

\textbf{\textbf{(ii)}} In the second case, if we apply HSIC with the same idea, we could compute HSIC of a hyperparameter with irrelevant values coming from configurations where it is not involved. Two situations can occur. First, if a conditional variable $\rx_i$ is never involved in the hyperparameter configurations that yield the $p$-percent best accuracies (depending on the percentile chosen), the values used for computing $S_{ \rx_i , \mathbf{Y}}$, i.e. $\rx_i\vert\rz=1$, are drawn from the initial, uniform distribution $\ru_i$. Then, $S_{ \rx_i , \mathbf{Y}}$ will be very low, and the conclusion will be that it is not impactful for reaching the percentile, which is correct since none of the best neural networks have used this hyperparameter. However, if $\rx_i$ is only involved in a subset of all tested hyperparameter configurations and is impactful in that case, $S_{ \rx_i , \mathbf{Y}}$ would be lowered by the presence of the other artificial values of $\rx_i$ drawn from the uniform distribution. In that case, we could miss its actual impact. The following example illustrates this phenomenon.\\

\begin{example}
\label{ex3} Let $f: [0,2]^3 \rightarrow \{0,1\}$ such that:

\begin{equation*}
       f(\rx_1, \rx_2, \rx_3) = 
   \begin{dcases}
    B & \text{if  } \rx_1 \in [0,1], \rx_2 \in [0,t]\\
    1 & \text{if  } \rx_1 \in [0,1], \rx_2 \in [t,2], \rx_3 \in [0,1],\\
    0 & \text{otherwise,  } 
    \end{dcases}
\end{equation*}
\end{example}

With $B$ a Bernoulli variable of parameter $0.5$ and $t\in [0,2]$ (so that $S_{\rx_2, \mathbf{Y}}$ is low). Let $\mathbf{Y}=\{1\}$. In that case, $\rx_1$ plays a key role for reaching $\mathbf{Y}$, and $\rx_3$ is taken into account only when $\rx_2 > t$. In these cases, it is as important as $\rx_1$ for reaching $\mathbf{Y}$ and we would like to retrieve this information. Parameter $t$ allows controlling how many values of $\rx_3$ will be involved. We evaluate $f$ on $n_s=2000$ points uniformly distributed across $[0,2]^3$, first with $t=1$. 

\begin{figure}[!ht]
   \begin{subfigure}{0.245\textwidth}
     \includegraphics[width=1.0\linewidth]{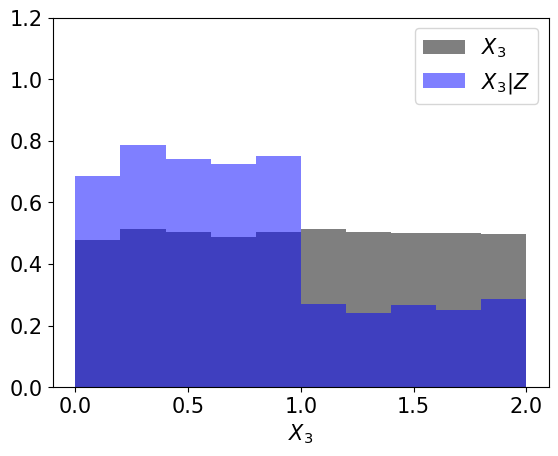}
     
   \caption{}
   \label{fig:cond_hista}
   \end{subfigure}
   \begin{subfigure}{0.245\textwidth}
     \includegraphics[width=1.0\linewidth]{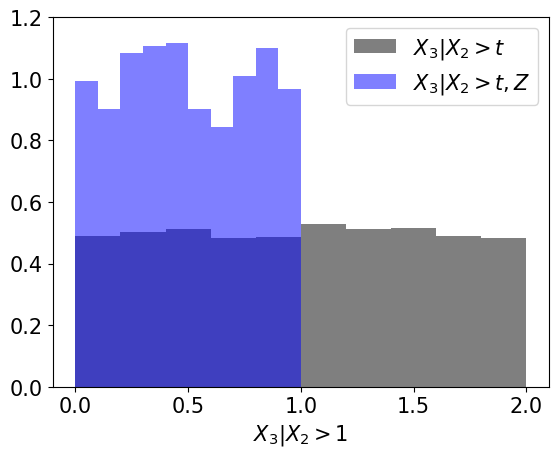}
     
    \caption{}
    \label{fig:cond_histb}
   \end{subfigure}
      \begin{subfigure}{0.245\textwidth}
     \includegraphics[width=1.0\linewidth]{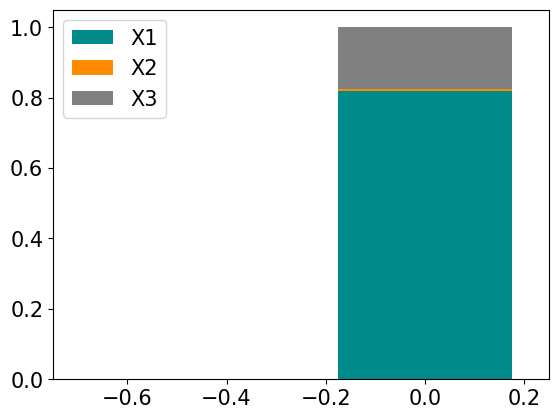}
     
    \caption{}
    \label{fig:cond_histc}
   \end{subfigure}
      \begin{subfigure}{0.245\textwidth}
     \includegraphics[width=1.0\linewidth]{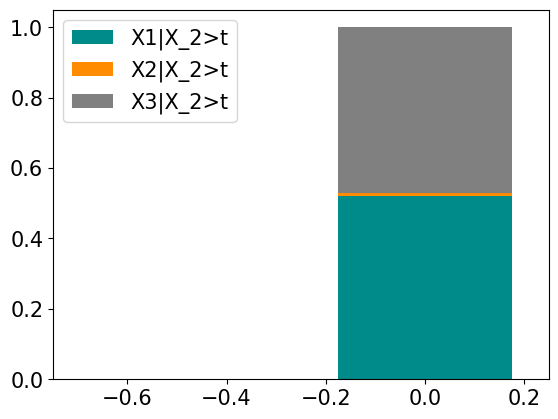}
     
    \caption{}
    \label{fig:cond_histd}
   \end{subfigure}
   \caption{ \textbf{(a)} - Histogram of $\rx_3$ and $\rx_3\vert\rz=1$ \textbf{(b)} - Histogram of $\rx_3$ and  $\rx_3\vert\rx_2 > t, \rz$.\\ \textbf{(c)} - $S_{\rx, \mathbf{Y}}$ for $\rx_1$; $\rx_2$ and $\rx_3$. \textbf{(d)} - $S_{\rx, \mathbf{Y}}$ for $\rx_1\vert\rx_2 > t$; $\rx_2\vert\rx_2 > t$ and $\rx_3\vert\rx_2 > t$.}
   
\end{figure}

\begin{figure}[!ht]
   \begin{subfigure}{0.02\textwidth}
        \rotatebox{90}{\textbf{(ii)}}
   \end{subfigure}
   \begin{subfigure}{0.235\textwidth}
     \includegraphics[width=1.0\linewidth]{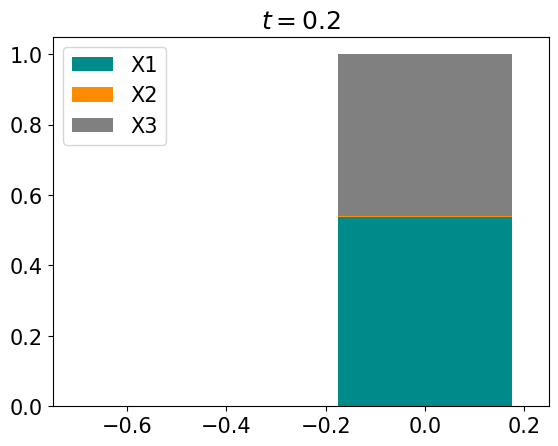}
     \end{subfigure}
    \begin{subfigure}{0.235\textwidth}
          \includegraphics[width=1.0\linewidth]{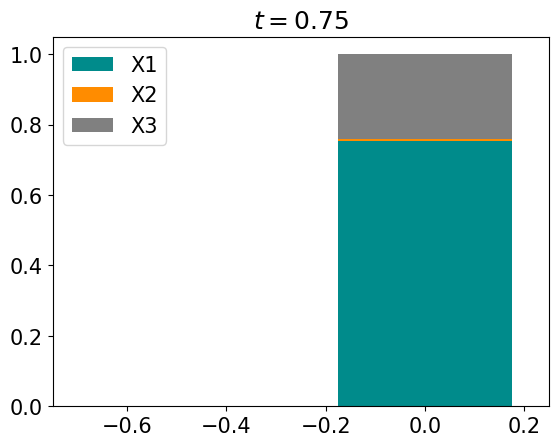}
   \end{subfigure}
   \begin{subfigure}{0.235\textwidth}
     \includegraphics[width=1.0\linewidth]{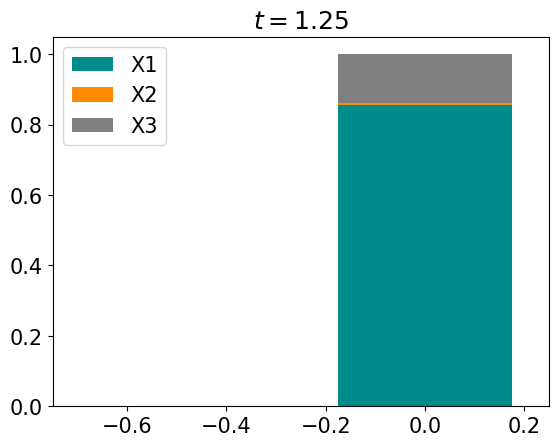}
     \end{subfigure}
     \begin{subfigure}{0.235\textwidth}
          \includegraphics[width=1.0\linewidth]{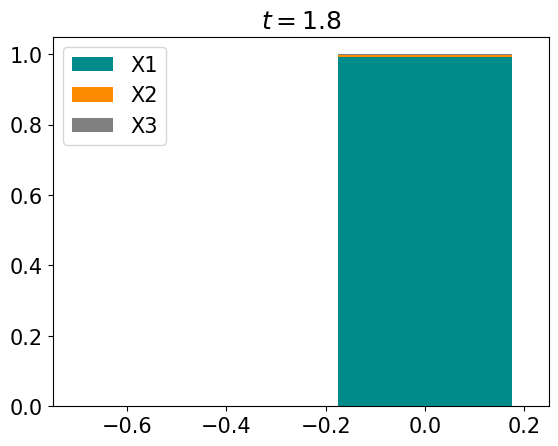}
   \end{subfigure}
       \begin{subfigure}{0.02\textwidth}
        \rotatebox{90}{\textbf{(iii)}}
   \end{subfigure}
      \begin{subfigure}{0.235\textwidth}
     \includegraphics[width=1.0\linewidth]{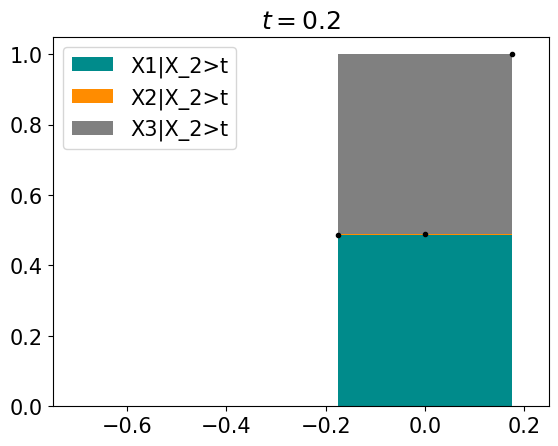}
     \end{subfigure}
     \begin{subfigure}{0.235\textwidth}
          \includegraphics[width=1.0\linewidth]{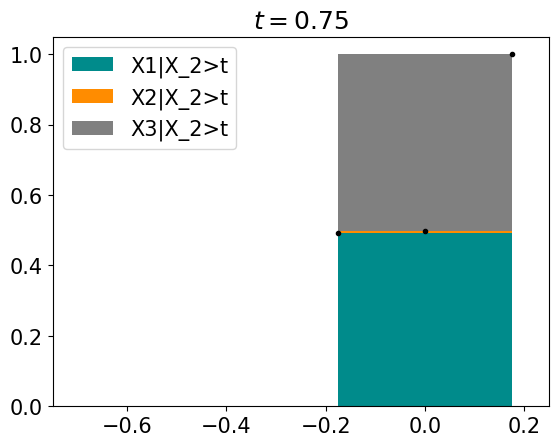}
   \end{subfigure}
      \begin{subfigure}{0.235\textwidth}
     \includegraphics[width=1.0\linewidth]{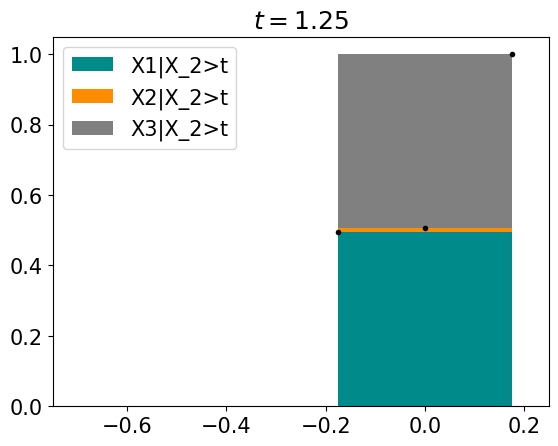}
     \end{subfigure}
     \begin{subfigure}{0.235\textwidth}
          \includegraphics[width=1.0\linewidth]{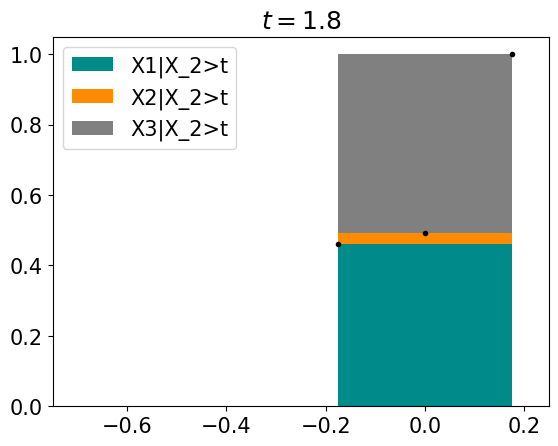}
   \end{subfigure}
   \caption{\textbf{Top (ii): }$S_{\rx,\mathbf{Y}}$ for $\rx_1$, $\rx_2$ and $\rx_3$ for different values of $t$.
   \textbf{Bottom (iii): }$S_{\rx,\mathbf{Y}}$ for $\rx_1\vert\rx_2> t$, $\rx_2\vert\rx_2> t$ and $\rx_3\vert\rx_2> t$ for different values of $t$.}
   \label{fig:cond_hist2}

\end{figure}

Figure \ref{fig:cond_hista} compares the histograms of $\rx_3$ and $\rx_3\vert\rz=1$. Figure \ref{fig:cond_histb} compares histograms of $\rx_3\vert\rx_2>t$ and of $\rx_3\vert\rx_2>t,\rz$. This shows that the distribution of $\rx_3\vert\rz=1$ is different if we choose to consider artificial values of $\rx_3$ or values of $\rx_3$ that are actually used by $f$ ($\rx_3\vert\rx_2 > t$). Figures \ref{fig:cond_histc} and \ref{fig:cond_histd} show that relative values of $S_{\rx_1, \mathbf{Y}}$ and $S_{\rx_3, \mathbf{Y}}$ are quite different whether we chose to consider $\rx_2 > t$ or not, meaning that the conclusions about the impact of $\rx_3$ can be potentially different. To emphasize how different these conclusions can be, we compare  $S_{\rx_1, \mathbf{Y}}$ and $S_{\rx_3, \mathbf{Y}}$ for different values of $t$. The results are displayed on Figure \ref{fig:cond_hist2} (top row). Since the value of $t$ controls how much artificial values there are for $\rx_3$, this demonstrates how different $S_{\rx_3, \mathbf{Y}}$ can be, depending on the amount of artificial points. This experiment emphasizes the problem because in all cases, $\rx_3$ is equally important for reaching $\mathbf{Y}$ whereas for $t=1.8$ we would be tempted to discard $\rx_3$.

To sum up, this formulation brings significant implementation advantages because it allows computing $S_{\rx_i, \mathbf{Y}}$ as if there were no conditionality. However, it carries a risk to miss essential impacts of conditional hyperparameters and discard them illegitimately. \\

\begin{figure}[!ht]
\centering
      \begin{subfigure}{0.4\textwidth}
     \includegraphics[width=1.0\linewidth]{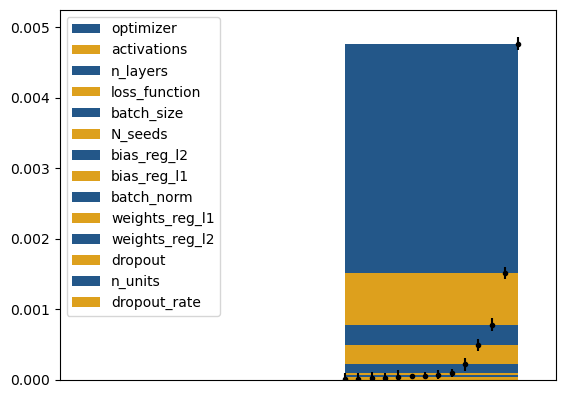}
     
     \caption{}
   \end{subfigure}
  \begin{subfigure}{0.4\textwidth}
     \includegraphics[width=1.0\linewidth]{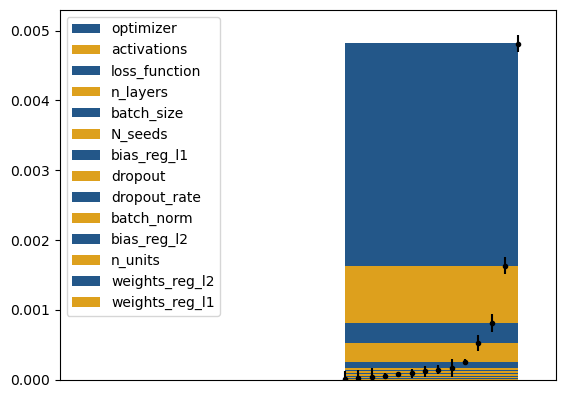}
     
     \caption{}
   \end{subfigure}
   \caption{Comparison of $S_{ \rx_i , \mathbf{Y}}$ for variables $\rx_i \in \mathcal{G}_0$ (a) and for variables $\rx_i \in \mathcal{G}_{\texttt{dropout\_rate}}$ (b)}
   \label{fig:hsicond}
\end{figure}

\textbf{\textbf{(iii)}}\label{sect:tree}
 In this work, we propose a splitting strategy that produces sets of hyperparameters that are involved together in the training, but are not disjoints, unlike \textbf{(i)}. Let $\mathcal{J}_k \in \{1,...,n_h\}$ be the set of indices of hyperparameters that can be involved in a training jointly with conditional hyperparameter $\rx_k$. We define $\mathcal{G}_{\rx_k} = \{\rx_i\vert\rx_k,  i \in \mathcal{J}_k\}$, the set of hyperparameters involved jointly in hyperparameter configurations when $\rx_k$ is also involved. By convention, we denote the set of all main hyperparameters by $\mathcal{G}_0$. In Runge problem, \texttt{dropout\_rate} is the only conditional hyperparameter, so we have two sets $\mathcal{G}_0 = \{\rx_1,...,\rx_{n_h}\} \setminus \texttt{dropout\_rate}$ and $\mathcal{G}_{\texttt{dropout\_rate}} = \{\rx_1\vert\texttt{dropout\_rate},...,\rx_{n_h}\vert\texttt{dropout\_rate}\}$ $= \{\rx_1\vert\texttt{dropout} = \texttt{true},...,\rx_{n_h}\vert\texttt{dropout} = \texttt{true}\}$. It is then possible to compute $S_{\rx_i,\mathbf{Y}}$ for $\rx_i \in \mathcal{G}_0$, identify the most impactful main hyperparameters, then to compute $S_{\rx_i,\mathbf{Y}}$ for $\rx_i \in \mathcal{G}_{\texttt{dropout\_rate}}$ and to assess if \texttt{dropout\_rate} is impactful by comparing it to other variables of $\mathcal{G}_{\texttt{dropout\_rate}}$.
On the example problem, we can compute $S_{\rx_i, \mathbf{Y}}$ only for $\rx_1$, $\rx_2$ and $\rx_3$ when $\rx_2> t$. This set would be $\mathcal{G}_{\rx_3}$ (except that $\rx_2$ is not categorical nor integer - but in that case we can consider $\bar{X}_2 = \1(\rx_2 > t)$). On the bottom row of Figure \ref{fig:cond_hist2}, $S_{\rx_1\vert\rx_2 > t, \mathbf{Y}}$ and $S_{\rx_3\vert\rx_2 > t, \mathbf{Y}}$ keep approximately the same values for all $t$, which is the correct conclusion since when $\rx_3$ is involved (i.e. $\rx_2 > t$) , it is as important as $\rx_1$ for reaching $\mathbf{Y}$. Coming back to Runge, Figure \ref{fig:hsicond} displays $S_{\rx_i,\mathbf{Y}}$ for Runge approximation for $\rx_i \in \mathcal{G}_{\texttt{dropout\_rate}}$, compared to the first approach where we do not care about conditionality, though in this specific case it does not change much of the conclusion that \texttt{dropout\_rate} is not impactful.

In Runge example, we have only considered one conditional hyperparameter, which is \texttt{dropout\_rate}, leading to only two groups $\mathcal{G}_0$ and $\mathcal{G}_{\texttt{dropout\_rate}}$. For another, more complex example, we could introduce additional conditional hyperparameters such as SGD's \texttt{momentum}. In that case, there would be two additional groups. The group $\mathcal{G}_{\texttt{momentum}}$, that contains hyperparameters conditioned to when \texttt{momentum} is involved, but also $\mathcal{G}_{(\texttt{dropout\_rate},\texttt{momentum})}$ that contains hyperparameters conditioned to when \texttt{momentum} and \texttt{dropout\_rate} are \textit{simultaneously} involved. If the initial random search contains $n_s$ configurations, \texttt{dropout\_rate} and \texttt{momentum} are involved in $n_s/2$ configurations. HSIC estimation of hyperparameters of the groups $\mathcal{G}_{\texttt{dropout\_rate}}$ and $\mathcal{G}_{\texttt{momentum}}$ will be coarser but still acceptable. However,  \texttt{dropout\_rate} and \texttt{momentum} would only be involved simultaneously in $n_s/4$ configurations, which may lead to too inaccurate HSIC estimation for $\mathcal{G}_{(\texttt{dropout\_rate},\texttt{momentum})}$. This happens because \texttt{dropout\_rate} and \texttt{momentum} do not depend on the same main hyperparameter. Hence, to avoid this problem, we only consider groups $\mathcal{G}$ with conditional hyperparameters that depend on the same main hyperparameter. In our case, these groups are $\mathcal{G}_0$, $\mathcal{G}_{\texttt{dropout\_rate}}$ and $\mathcal{G}_{\texttt{momentum}}$.

\subsection{Summary: evaluation of HSIC in hyperparameter analysis}

In this section, we summarize the results of the previous discussions to provide a methodology for evaluating the HSIC of hyperparameters in complex search spaces in Algorithm \ref{alg:hsic}.\\

\textbf{Comments on Algorithm \ref{alg:hsic}.} \textbf{Line 1:} one can choose any initial distribution for hyperparameters. \textbf{Line 2:} this step is a classical random search. Recall that HSIC evaluation can be applied after any random search, even if it was not initially conducted for HSIC estimation. Configurations $\boldsymbol{\sigma}_i$ are sampled from $\boldsymbol{\sigma} = (\rx_1, ..., \rx_{n_h}) \in \mathcal{H}$. \textbf{Line 3: } this step strongly benefits from parallelism. \textbf{Line 4: } the set $\mathbf{Y}$ is often taken as the $p$ \% percentile of $\{Y_1,...,Y_{n_s}\}$, but can be any other set depending on what we want to assess. \textbf{Line 6 - 10:} the evaluation starts with main hyperparameters because they are always involved. Once most impactful main hyperparameters are selected, we assess the conditional ones.  

\begin{algorithm}[h!]
   \caption{ \small Evaluation of HSIC in hyperparameter analysis}
   \label{alg:hsic}
    \begin{algorithmic}[1]
   \State {\bfseries Inputs: } hyperparameter search space $\mathcal{H} = \mathcal{X}_1 \times ... \times \mathcal{X}_{n_h}$, $n_s$.
   \State Sample $n_s$ hyperparameter configurations $\{\boldsymbol{\sigma}_1,..., \boldsymbol{\sigma}_{n_s}\}$.
   \State Train a neural network for each configuration and gather outputs $\{Y_1,...,Y_{n_s}\}$.
   \State Define $\mathbf{Y}$. 
   \State Construct conditional groups $\mathcal{G}_0,...$.
   \For{each group, starting with $\mathcal{G}_0$ }
       \State Construct $\ru_i$ for every $\rx_i$ using $\Phi_i$ of section \ref{normalization}.
       \State Compute $S_{\rx_i,\mathbf{Y}} := S_{\ru_i,\mathbf{Y}}$ using \eqref{eq:sxy}.
       \State By comparing them, select the most impactful hyperparameters.
       \State Check for interacting hyperparameters.
   \EndFor
   \State {\bfseries Outputs:} Most impactful hyperparameters and interacting hyperparameters. 
    \end{algorithmic}
    
\end{algorithm}

\begin{remark}
    The value of $S_{\rx_i,\mathbf{Y}}$ strongly depends on the initial distribution chosen for $\rx_i$. Indeed, if the distribution only spans values of $\rx_i$ that yield good prediction error,  $S_{\rx_i,\mathbf{Y}}$ will be low. Conversely, if it spans good values but also includes absurd values, $S_{\rx_i,\mathbf{Y}}$ will be higher. Hence, without \textit{a priori} knowledge, we recommend to select a large range of values for each $\rx_i$
\end{remark}

\section{Experiments}
\label{sect:usages}

\label{sect:hsic}

Now that we can compute and correctly assess HSIC, we introduce possible usages of this metric in the context of hyperparameter analysis. In this section, we explore three benefits that we can draw from HSIC based hyperparameter analysis.
 \begin{itemize}
     \item Interpretability: HSIC allows analyzing hyperparameters, obtaining knowledge about their relative impact on error. 
     \item Stability: Some hyperparameter configurations can lead to dramatically high errors. A hyperparameters range reduction based on HSIC can prevent such situations.
     \item Acceleration: We can choose values for less important hyperparameters that improve inference and training time.
     
 \end{itemize}
 
 We illustrate these points through hyperparameter analysis when training a fully connected neural network on MNIST and a convolutional neural network on Cifar10. We also study the approximation by a fully connected neural network of Bateman equations solution. Details about the construction of Bateman equations data set can be found in \textbf{\hyperref[appC]{Appendix C}} and hyperparameter spaces and conditional groups $\mathcal{G}_0,...$ for each problem in \textbf{\hyperref[appA]{Appendix A}}.
 
 \subsection{Hyperparameter analysis}

This section presents a first analysis of the estimated value of HSIC for the three benchmark data sets: MNIST, Cifar10, and Bateman equations. These evaluations are based on an initial random search for $n_s=1000$ different hyperparameter configurations.  The set $\mathbf{Y}$ is the $10\%$-best errors percentile, so $n_s$ is taken sufficiently large for HSIC to be correctly estimated. Indeed, if $n_s=1000$, there will be $100$ samples of $\ru_i\vert\rz=1$. For every data set, we extract 10\% of the training data to construct a validation set to evaluate $\rz$. We keep a test set for evaluating neural networks obtained after hyperparameter optimization described in Section \ref{sect:tso}. This random search was conducted using $100$ parallel jobs on CPU nodes for fully connected neural networks and 24 parallel jobs on Nvidia Tesla V100 and P100 GPUs for convolutional neural networks, so the results for these configurations were obtained quite quickly, in less than two days.

Note that for each data set, graphical comparison of $S_{ \rx_i , \mathbf{Y}}$ for conditional groups $\mathcal{G}_0,...$ is displayed in \textbf{\hyperref[appB]{Appendix B}}, for conciseness and clarity.

\subsubsection{MNIST}

\begin{figure}[!ht]
\centering

      \begin{subfigure}{0.32\textwidth}
     \includegraphics[width=1.0\linewidth]{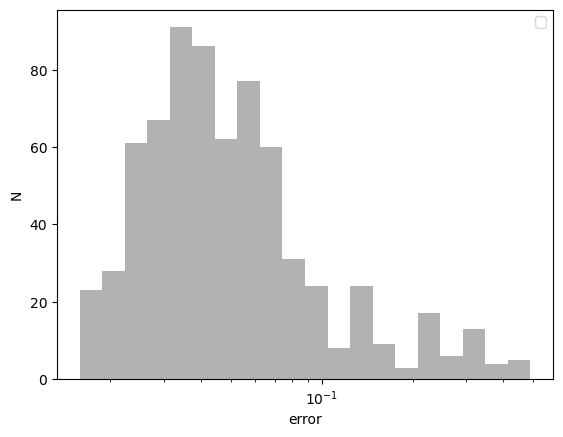}
     
   \end{subfigure}
      \begin{subfigure}{0.32\textwidth}
     \includegraphics[width=1.0\linewidth]{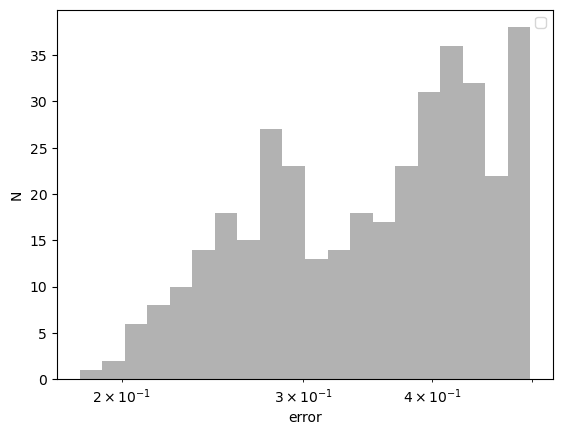}
     
   \end{subfigure}
         \begin{subfigure}{0.32\textwidth}
     \includegraphics[width=1.0\linewidth]{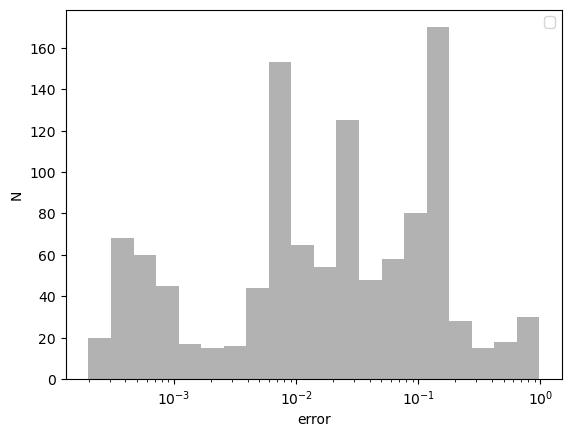}
     
   \end{subfigure}
     \begin{subfigure}{0.32\textwidth}
     \includegraphics[width=1.0\linewidth]{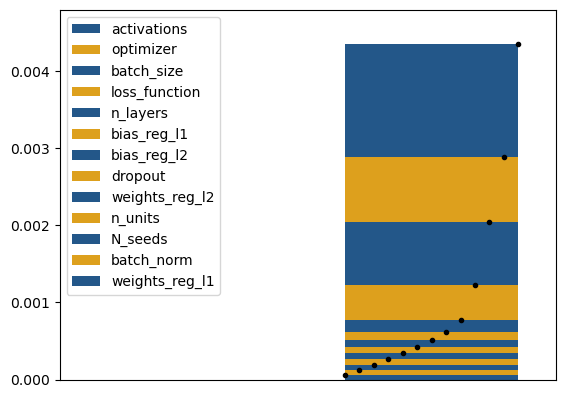}
     \caption{MNIST}
    \label{fig:hsicmnist}
   \end{subfigure}
         \begin{subfigure}{0.32\textwidth}
     \includegraphics[width=1.0\linewidth]{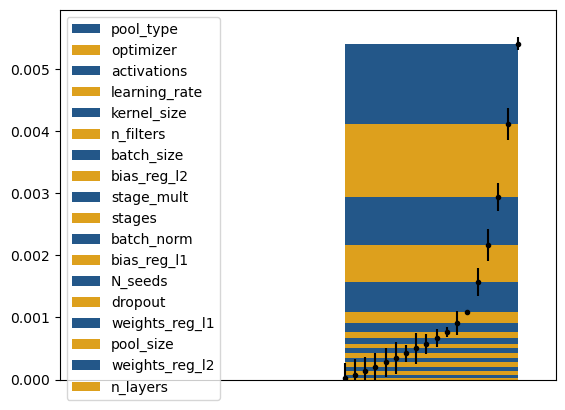}
     \caption{Cifar10}
        \label{fig:hsiccifar}
     \end{subfigure}
  \begin{subfigure}{0.32\textwidth}
     \includegraphics[width=1.0\linewidth]{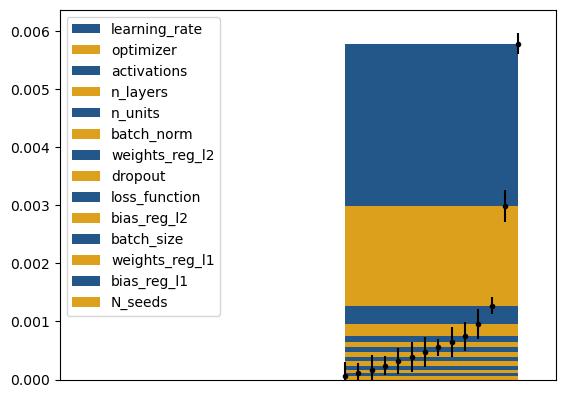}
     \caption{Bateman}
        \label{fig:hsicbateman}
     \end{subfigure}
    \caption{(top) Histograms of the initial random sampling of configurations and (bottom) comparison of $S_{\rx_i, \mathbf{Y}}$ for every main hyperparameters. }

\end{figure}

We train $n_s=1000$ different neural networks. We can see on Figure \ref{fig:hsicmnist} that the accuracy goes up to $\sim99\% (1 - \text{error})$ which is quite high for a fully connected neural network on MNIST. Figure \ref{fig:hsicmnist} also displays the values of $S_{ \rx_i , \mathbf{Y}}$ for each hyperparameter $\rx_i$ stacked vertically. Here, \texttt{activation}, \texttt{optimizer}, \texttt{batch\_size} and \texttt{loss\_function} have significantly high $S_{ \rx_i , \mathbf{Y}}$. Hyperparameter \texttt{n\_layers} also stands out from the remaining hyperparameter, while staying far below \texttt{loss\_function} HSIC. There is one conditional group to consider, $\mathcal{G}_{\texttt{dropout\_rate}}$, and \texttt{dropout\_rate} is found not to be impactful. 

Interestingly, neither the depth (\texttt{n\_layers}) nor the width (\texttt{n\_units}) are among the most important hyperparameters. Notice that the random search yields a neural network of depth $4$ and width $340$ which obtained $98.70 \%$ accuracy, while the best networks (there were two) obtained $98.82\%$ accuracy for a depth of $10$ and a width of $791$ and $1403$, respectively. Recall that the min-max depth was $1$-$10$ and width was $134$-$1500$. It means that lighter networks are capable of obtaining competitive accuracy. Another interesting observation is that \texttt{loss\_function} does not have the highest HSIC, meaning that Mean Squared Error allows obtaining good test errors, which is surprising for a classification problem.

\begin{figure}[!t]
\centering
      \begin{subfigure}{0.32\textwidth}
     \includegraphics[width=1.0\linewidth, trim={0.5cm 0cm 0cm 1cm}]{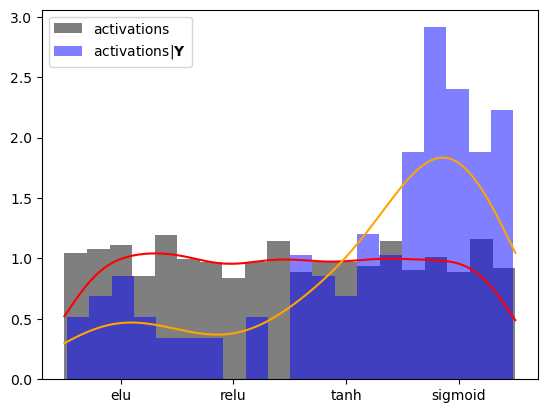}
     
   \end{subfigure}
      \begin{subfigure}{0.32\textwidth}
     \includegraphics[width=1.0\linewidth, trim={0.5cm 0cm 0cm 1cm}]{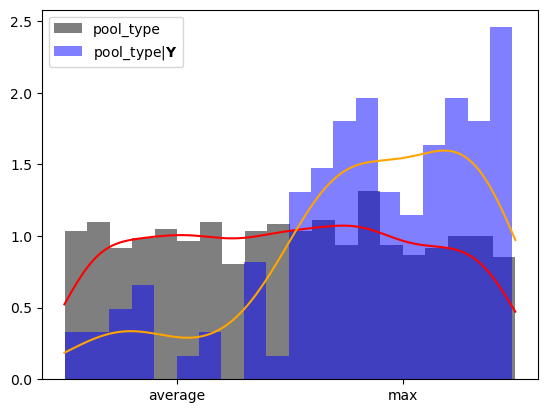}
     
   \end{subfigure}
         \begin{subfigure}{0.32\textwidth}
     \includegraphics[width=1.0\linewidth, trim={0.5cm 0cm 0cm 1cm}]{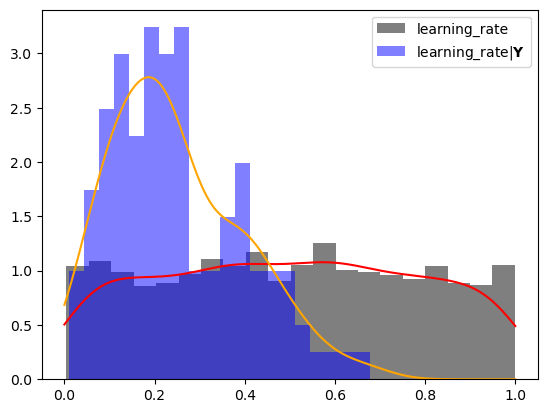}
     
   \end{subfigure}
     \begin{subfigure}{0.32\textwidth}
     \includegraphics[width=1.0\linewidth, trim={0.5cm 0cm 0cm 1cm}]{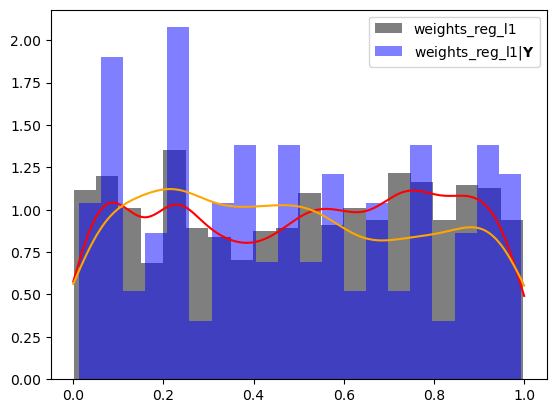}
     \caption{MNIST}
     \label{fig:kdemnist}
   \end{subfigure}
         \begin{subfigure}{0.32\textwidth}
     \includegraphics[width=1.0\linewidth, trim={0.5cm 0cm 0cm 1cm}]{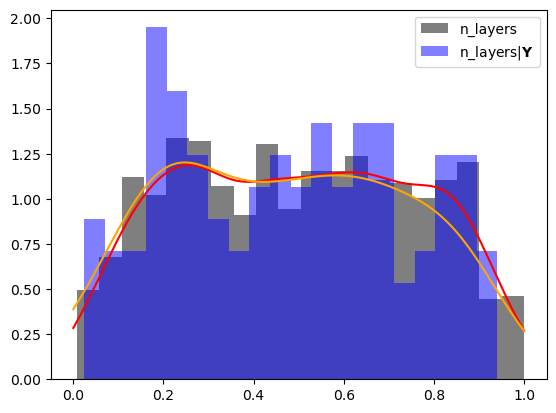}
     \caption{Cifar10}
     \label{fig:kdecifar}
     \end{subfigure}
  \begin{subfigure}{0.32\textwidth}
     \includegraphics[width=1.0\linewidth, trim={0.5cm 0cm 0cm 1cm}]{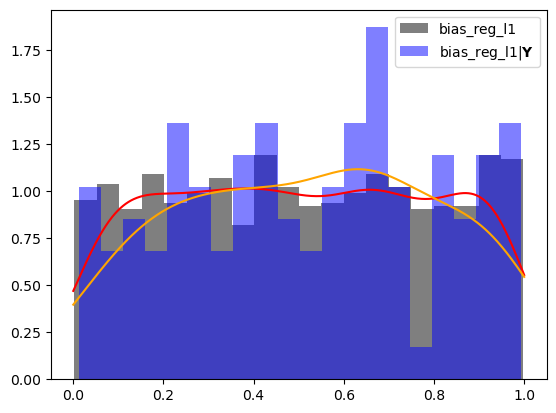}
     \caption{Bateman}
     \label{fig:kdebateman}
     \end{subfigure}
    \caption{Representation of $\ru_i\vert\rz=1$ (orange for KDE and blue for histogram) and $\ru_i$ (red for KDE and grey for histogram), for hyperparameters $\rx_i$ with high (top) and low (bottom) $S_{\rx_i, \mathbf{Y}}$}
   
\end{figure}

We plot histograms of $\ru_i$ and $\ru_i\vert\rz=1$ on Figure \ref{fig:kdemnist} for \texttt{activation} (top) and \texttt{weights\_reg\_l1} (bottom) with repeated sampling for categorical hyperparameters, like in Section \ref{normalization}. Note that the first and the second hyperparameters have respectively a high and low $S_{\rx_i, \mathbf{Y}}$. We can see that for hyperparameters with high $S_{\rx_i, \mathbf{Y}}$, $\ru_i\vert\rz=1$ (orange for KDE, blue for histogram) is quite different from $\ru_i$ (red for KDE, gray for histogram). On the contrary, for hyperparameters with low $S_{\rx_i, \mathbf{Y}}$ there not seems to have major differences.  

\subsubsection{Cifar10}

We train $n_s=1000$ different convolutional neural networks. After the initial random search, the best validation error is $81.37 \%$. Note that the histogram of Figure \ref{fig:hsiccifar} is truncated because many hyperparameter configurations led to diverging errors. Here, \texttt{pool\_type}, \texttt{optimizer}, \texttt{activation}, \texttt{learning\_rate} and \texttt{kernel\_size} have the highest $S_{\rx_i, \mathbf{Y}}$, followed by \texttt{n\_filters}. Half of these hyperparameters are specific to convolutional neural networks, which validates the impact of these layers on classification tasks for image data. The conditional groups are listed in \textbf{\hyperref[appA]{Appendix A}}. We do not show $S_{\rx_i, \mathbf{Y}}$ comparisons for every group for clarity of the article and simply report that one conditional hyperparameter \texttt{centered}, which triggers centered RMSprop if this value is chosen for \texttt{optimizer}, is also found to be impactful.

The depth (\texttt{n\_layers}) is the less important hyperparameters. Here, the random search yields a neural network of depth $4$ and width $53$, with $3$ stages (meaning that the neural network is widened $3$ times), which obtained $80.70 \%$ validation accuracy, while the best networks obtained $81.37\%$ accuracy for a depth of $6$ and $48$ but $4$ stages. The conclusion is the same as for MNIST: increasing the size of the network is not the only efficient way to improve its accuracy.

We plot histograms of $\ru_i$ and $\ru_i\vert\rz=1$ on Figure \ref{fig:kdecifar} for \texttt{pool\_type} (top) and \texttt{n\_layers} (bottom) like in the previous section. The histograms of \texttt{n\_layers} are interesting because even the histogram of $\ru_i$ does not seem uniform. An explanation could be that configurations lead to out-of-memory errors or are so long to train that $1000$ other neural networks with different configurations have already been trained meanwhile. It also explains why its HSIC is so low. Still, the conclusions that \texttt{n\_layers} has a limited impact is valid since there is no major differences between $\ru_i$ and $\ru_i\vert\rz=1$.

\subsubsection{Bateman equations}

For Bateman equations, mean squared error goes down to $2.90 \times10^{-5}$. Like for Cifar10, the histogram of Figure \ref{fig:hsiccifar} is truncated because many hyperparameter configurations led to diverging errors. For this problem, \texttt{learning\_rate}, \texttt{optimizer}, \texttt{activations} and \texttt{n\_layer} can be considered as impactful. Conditional groups are also listed in \textbf{\hyperref[appA]{Appendix A}}. Three conditional hyperparameters are important: \texttt{beta\_2}, the second moment decay coefficient of Adam and Nadam, \texttt{nesterov}, that triggers Nesterov's momentum in SGD and \texttt{centered}, described previously.

HSIC for \texttt{n\_layers} is still the lowest of the significant $S_{\rx_i, \mathbf{Y}}$ and  \texttt{n\_units} belongs to less impactful hyperparameters. We perform the same analysis as for MNIST and Cifar10 and quote that the best neural network has depth $5$ and width $470$ while another neural network of depth $5$ and width $62$ reaches $3.74 \times 10 ^{-5}$ validation error. 

We plot histograms of $\ru_i$ and $\ru_i\vert\rz=1$ on Figure \ref{fig:kdebateman} for \texttt{learning\_rate} (top) and \texttt{bias\_reg\_l1} (bottom). Histograms of \texttt{learning\_rate} is interesting because this hyperparameter is continuous so the distribution $\ru_i\vert\rz=1$ seems more natural. This once again illustrates the differences of $\ru_i$ and $\ru_i\vert\rz=1$ for hyperparameters with high and low  $S_{\rx_i, \mathbf{Y}}$.

\subsection{Modification of hyperparameters distribution to improve training stability}\label{sect:reduction_d}

Up to now, we only considered $\mathbf{Y}$ to be the $10\%$ best error percentile, which is natural since we want to understand the impact of hyperparameters towards good errors. However, HSIC formalism and our adaptation to hyperparameter analysis allow us to choose any $\mathbf{Y}$. In the previous section, for Cifar10 and Bateman, we truncated histograms of Figure \ref{fig:hsiccifar} because many hyperparameter configurations led to diverging errors. It is possible to understand why by choosing $\mathbf{Y}$ as the set of the $10\%$ worst errors. Then, HSIC can be applied to assess the importance of each hyperparameter towards the worst errors.

\begin{figure}[!ht]
\centering
   \begin{subfigure}{0.35\textwidth}
     \includegraphics[width=1.0\linewidth]{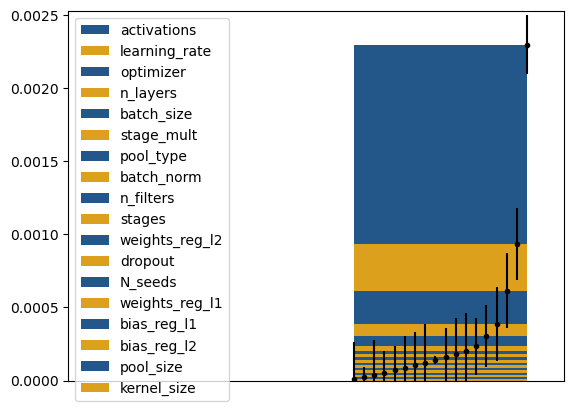}
     
   \end{subfigure}
   \begin{subfigure}{0.3\textwidth}
     \includegraphics[width=1.0\linewidth]{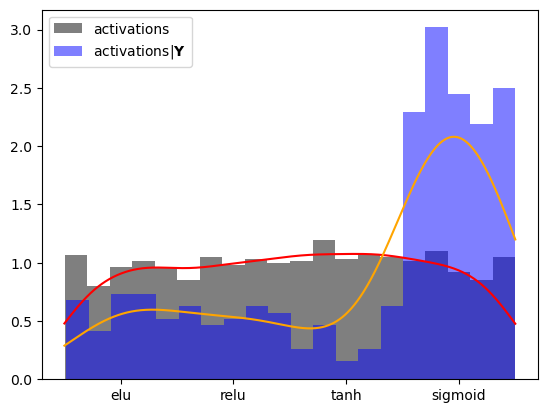}
     
   \end{subfigure}
      \begin{subfigure}{0.3\textwidth}
     \includegraphics[width=1.0\linewidth]{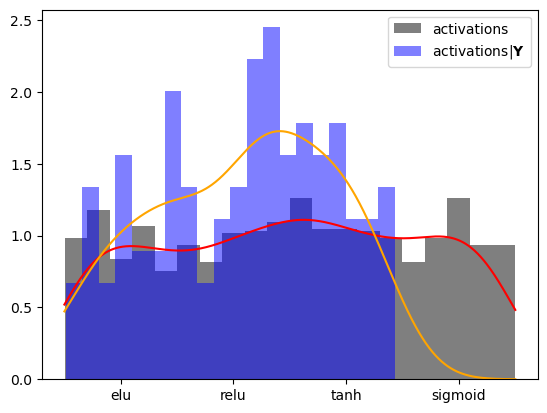}
     
   \end{subfigure}
      \begin{subfigure}{0.35\textwidth}
     \includegraphics[width=1.0\linewidth]{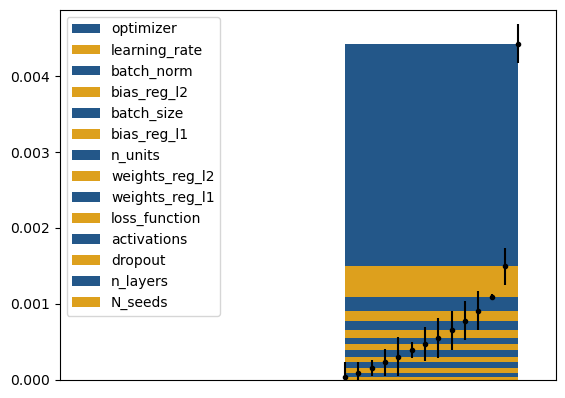}
     
     \caption{$S_{ \rx_i , \mathbf{Y}}$, $\mathbf{Y} = 10\%$ worst}
     \label{fig:appCa}
   \end{subfigure}
  \begin{subfigure}{0.3\textwidth}
     \includegraphics[width=1.0\linewidth]{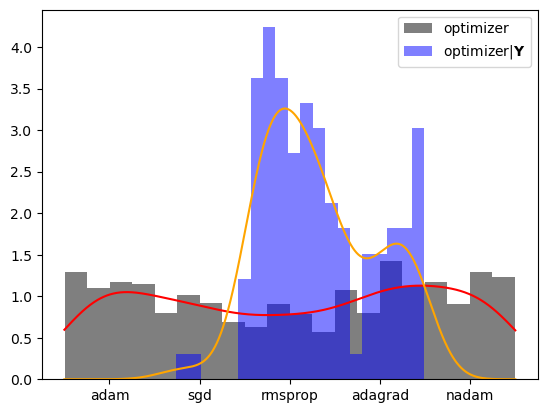}
     
     \caption{$\rx_i\vert\mathbf{Y}$, $\mathbf{Y} = 10\%$ worst}
     \label{fig:appCb}
   \end{subfigure}
    \begin{subfigure}{0.3\textwidth}
     \includegraphics[width=1.0\linewidth]{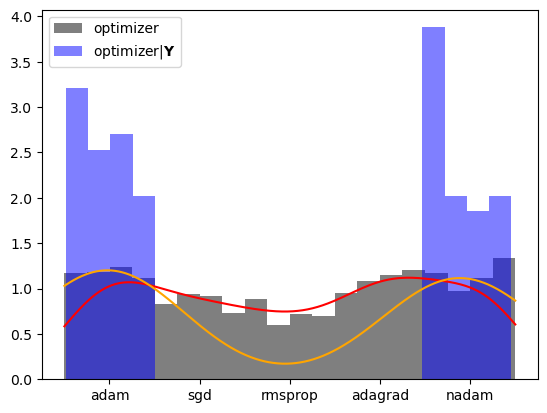}
     
     \caption{$\rx_i\vert\mathbf{Y}$, $\mathbf{Y}=10\%$ best}
     \label{fig:appCc}
   \end{subfigure}
   \caption{\textbf{Top}: Cifar10. \textbf{Bottom:} Bateman. \textbf{(a)} Comparison of $S_{ \rx_i , \mathbf{Y}}$ when $\mathbf{Y}$ is the set of the $10\%$ worst errors. \textbf{(b)} Histogram of  $\rx_i\vert\mathbf{Y}$ when $\mathbf{Y}$ is the set of  $10\%$ worst errors, with $\rx_i=\texttt{activations}$ for Cifar10 and $\rx_i=\texttt{optimizer}$ for Bateman. \textbf{(c)} Histogram of  $\rx_i\vert\mathbf{Y}$ when $\mathbf{Y}$ is the set of the $10\%$ best errors, with $\rx_i=\texttt{activations}$ for Cifar10 and $\rx_i=\texttt{optimizer}$ for Bateman.}
   
\end{figure}

Figure \ref{fig:appCb} shows $S_{ \rx_i , \mathbf{Y}}$ comparisons, for Cifar10 and Bateman, when  $\mathbf{Y}$ is the set of the $10\%$ worst errors. In that case, $S_{ \rx_i , \mathbf{Y}}$ measures how detrimental bad values of $\rx_i$ can be for the neural network error. For Cifar10, \texttt{activation} is the main responsible for the highest errors. If we plot the histogram of $\texttt{activation}\vert\mathbf{Y}$, we can see that \texttt{sigmoid} is a bad value in the sense that most of the worst neural networks use this activation function. If we come back to $\mathbf{Y}$ being the set of the $10\%$ best neural networks, we see that none of the best neural networks have \texttt{sigmoid} as the activation function. By itself, this kind of knowledge is valuable because it gives insights about hyperparameter's impact. It also directly brings some practical benefits: in that case, we could reasonably discard \texttt{sigmoid} from the hyperparameter space and therefore adapt the distribution of \texttt{activation} to improve stability. The same reasoning can be applied to Bateman, with $\rx_i=\texttt{optimizer}$, for \texttt{adagrad} and \texttt{rmsprop} optimizers.

 Note that we could have drawn the previous conclusions by directly looking at histograms as represented in Figure \ref{fig:appCb} and \ref{fig:appCc}. However, when the number of hyperparameters grows, the number of histograms to look at and to visually evaluate grows as well, and the analysis becomes tedious. Thanks to HSIC, we know directly which histograms to look at and how to rank hyperparameters when it is not visually clear-cut.

\subsection{Interval reduction for continuous or integer hyperparameters that affect execution speed}\label{sect:reduction}

One common conclusion of $S_{\rx_i, \mathbf{Y}}$ values for the last three machine learning problems is that one does not have to set high values for hyperparameters that affect execution speed, such as \texttt{n\_units}, \texttt{n\_layers}, or \texttt{n\_filters}, in order to obtain competitive models. It naturally raises the question of how to bias the hyperparameter optimization towards such models. Multi-objective hyperparameter optimization algorithms have already been successfully applied, like in \cite{hsic:mnasnet} for instance, but these algorithms are black-boxes and involve tuning additional hyperparameters for the multi-objective loss function. 

In our case, we can use information from $S_{\rx_i, \mathbf{Y}}$ to reduce the hyperparameter space search in order to obtain more cost-effective neural networks. The most simple way to achieve that goal is to select values that improve execution speed for hyperparameters which have low $S_{\rx_i, \mathbf{Y}}$ values. For MNIST, it would mean for instance to choose $\texttt{n\_units} = 128$, for Cifar10, $\texttt{n\_layers} = 3$ or for Bateman, $\texttt{n\_units} = 32$.

However, if all hyperparameters that affect execution speed are important, i.e. they have high $S_{\rx_i, \mathbf{Y}}$ value, we may not be able to apply the previous idea. In that case, we can use HSIC in a different way to still achieve our goal, for integer or continuous hyperparameters (such as \texttt{n\_layers}, \texttt{n\_units}, or \texttt{kernel\_size}). Note that most of the time, for these hyperparameters, a too low or high value will increase the error or the execution speed, respectively. We would like to choose a value which is as low as possible without hurting the error too much. Suppose that $\rx_i = \texttt{n\_layers} \in \{a,...,b\}$ and that $S_{\rx_i, \mathbf{Y}}$ is high, so that \texttt{n\_layers} is among the most important hyperparameters. It is likely that $S_{\rx_i, \mathbf{Y}}$ is high because $a$ is too small. One could therefore compute $S_{\rx_i\vert\rx_i \in \{a + c,...,b\}, \mathbf{Y}}$ for $c \in \{1, ...,b-a\}$, starting with $c=1$ until $S_{\rx_i\vert\rx_i \in \{a + c,...,b\}, \mathbf{Y}}$ becomes low. Then, hyperparameter $\texttt{n\_layers}$ can be replaced by $\texttt{n\_layers}\vert\texttt{n\_layers}\in \{a + c,...,b\}$, which has a low HSIC, and whose value can hence be set to $a + c$.

\begin{figure}[!ht]
    \centering
         \includegraphics[width=0.5\linewidth, trim={0.5cm 0cm 0cm 0cm}]{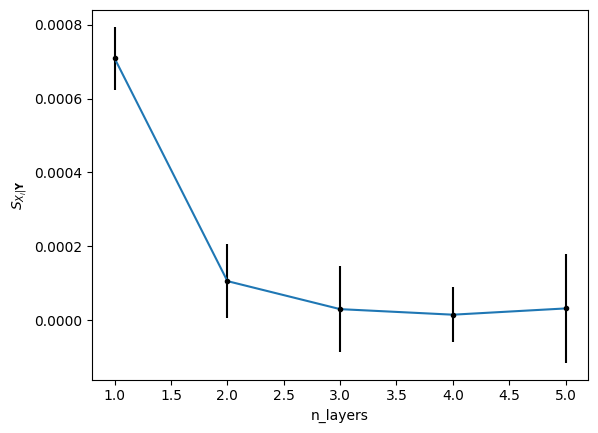}
    
        \caption{$S_{\rx_i\vert\rx_i \in \{a + c,...,b\}, \mathbf{Y}}$ w.r.t. $c$ for \texttt{n\_layers} in Runge. The error bars traduce the standard estimation error. }
        \label{fig:reduction0}
    \end{figure}

To illustrate this, let us come back to Runge data set. We first focus on this example because we have been able to train $n_s=10000$ different neural networks so the methodology can be tested with limited noise. In Figure \ref{fig:reduction0}, $S_{\rx_i\vert\rx_i \in \{a + c,...,b\}, \mathbf{Y}}$ is plotted with respect to $c$, where $\rx_i = \texttt{n\_layers}$. We see that $S_{\rx_i\vert\rx_i \in \{a + c,...,b\}, \mathbf{Y}}$ decreases until $\texttt{n\_layers} = 3$, after which the tendency is not statistically significant. Choosing $\texttt{n\_layers} = 3$ makes $\texttt{n\_layers}$ belong to the less important hyperparameters so it is a good trade-off value for execution speed and accuracy. 

\begin{figure}[!t]
    \centering
      \begin{subfigure}{0.32\textwidth}
        \includegraphics[width=1.0\linewidth, trim={0.5cm 0cm 0cm 0cm}]{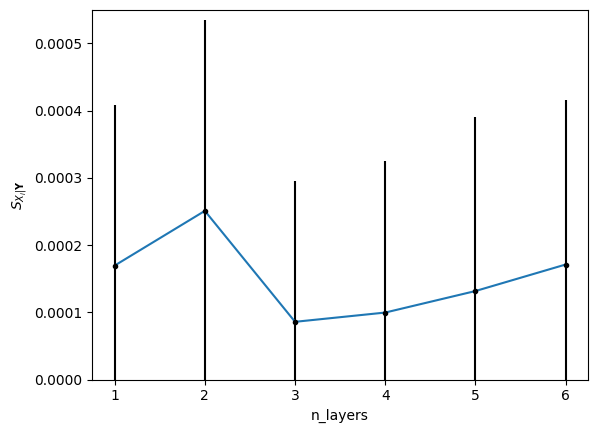}
      \end{subfigure}
            \begin{subfigure}{0.32\textwidth}
        \includegraphics[width=1.0\linewidth, trim={0.5cm 0cm 0cm 0cm}]{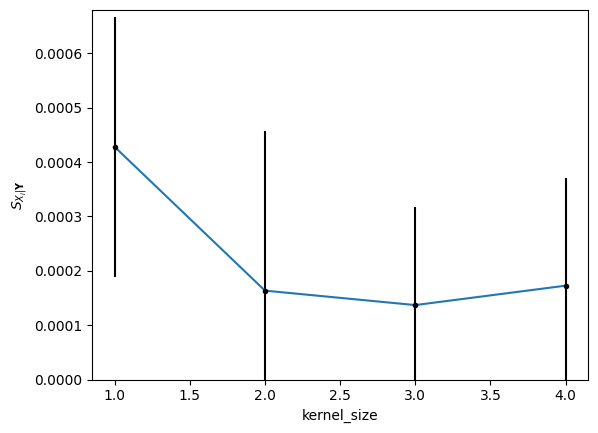}
        \end{subfigure}
      \begin{subfigure}{0.32\textwidth}
        \includegraphics[width=1.0\linewidth, trim={0.5cm 0cm 0cm 0cm}]{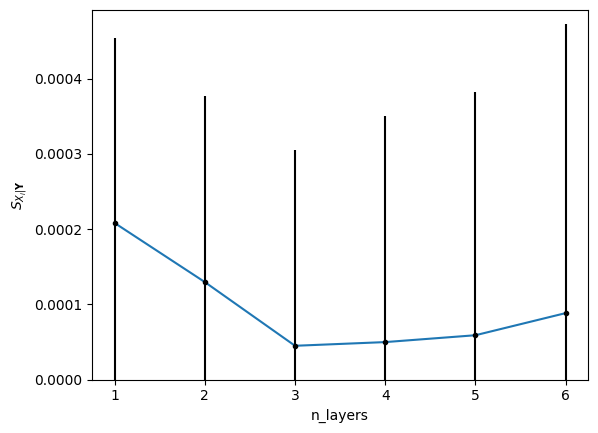}
        \end{subfigure}
         \begin{subfigure}{0.32\textwidth}
         \includegraphics[width=1.0\linewidth, trim={0.5cm 0cm 0cm 0cm}]{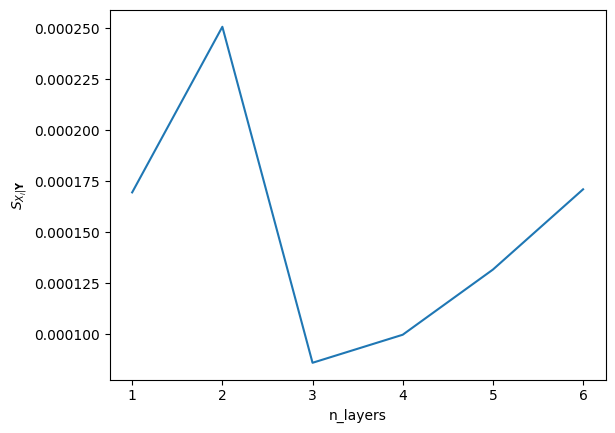}
         \caption{MNIST}
       \end{subfigure}
             \begin{subfigure}{0.32\textwidth}
         \includegraphics[width=1.0\linewidth, trim={0.5cm 0cm 0cm 0cm}]{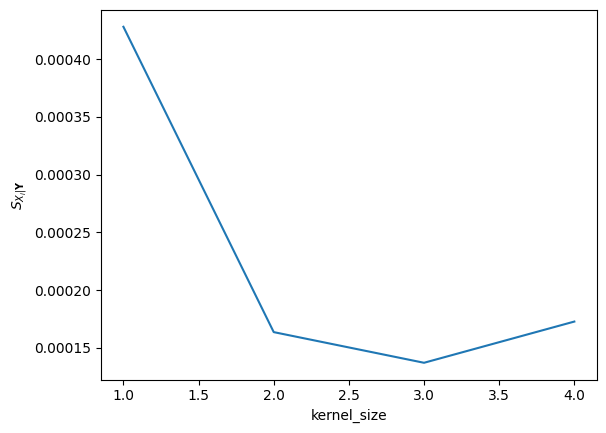}
         \caption{Cifar10}
         \end{subfigure}
      \begin{subfigure}{0.32\textwidth}
         \includegraphics[width=1.0\linewidth, trim={0.5cm 0cm 0cm 0cm}]{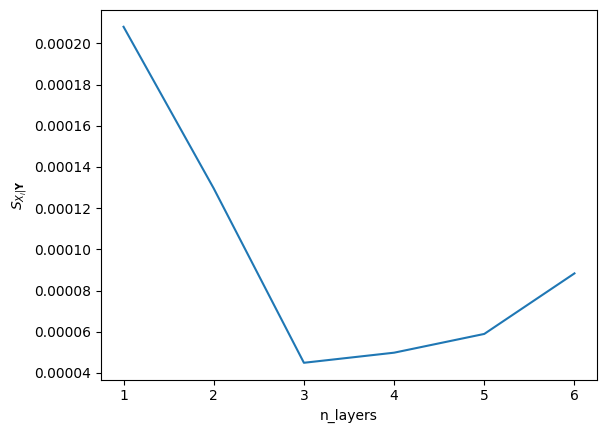}
         \caption{Bateman}
         \end{subfigure}
        \caption{$S_{\rx_i\vert\rx_i \in \{a + c,...,b\}, \mathbf{Y}}$ w.r.t. $c$ for (a) \texttt{n\_layers} in MNIST, (b) \texttt{kernel\_size} in Cifar10 and (c) \texttt{n\_layers} in Bateman with error bars (\textbf{top}) and without error bars (\textbf{bottom}). The error bars traduce the standard estimation error.}
        \label{fig:reduction}
    \end{figure}

We apply this methodology to MNIST, Cifar10, and Bateman problems in Figure \ref{fig:reduction}. When plotting these curves, too high values of $c$ have to be discarded since the more $c$ increases, the less points there are to compute $S_{\rx_i\vert\rx_i \in \{a + c,...,b\}, \mathbf{Y}}$. It could explain the strange behavior of the plots at the right of the axis of Figure \ref{fig:reduction}, and the widening of error bars for $\texttt{n\_layers}=5$ in Figure \ref{fig:reduction0}. 

Note that in Figure \ref{fig:reduction}, error bars are much larger than in Figure \ref{fig:reduction0}. Indeed, in these cases, $S_{\rx_i\vert\rx_i \in \{a + c,...,b\}, \mathbf{Y}}$ are evaluated with 10 times less points. Hence, one must be careful with their interpretation. First, we are far from the asymptotical regime under which estimation error is gaussian for so few estimation points. It explains why error bars can go below $0$ whereas the value to estimate is a distance. Therefore, these bars only indicate how spread the error is. Second, since it turns out that the error is very spread, the trade-off value must be chosen with caution by taking this statistic into account. In this manuscript, we rely on a human eye to qualitatively chose this value, but in future work, we should study the use of statistical tests. 

 Finally, these plots suggests that we could set $\texttt{n\_layers} = 3$ for MNIST, $\texttt{kernel\_size} = 3$ for Cifar10 and $\texttt{n\_layers} = 3$ for Bateman without affecting the error too much. Once the hyperparameter space has been reduced to improve neural networks execution time, it is possible to apply any classical hyperparameter optimization algorithm.

\section{Optimization by focusing on impactful hyperparameters}\label{sect:tso}

One of the most successful and widely used hyperparameter optimization algorithms is Gaussian Processes-based Bayesian Optimization, which we denote GPBO by convenience. However, this algorithm is known to struggle in too high dimensions. In the case of Cifar10, choosing values for hyperparameters that affect execution time would still lead to a space of dimension $20$, which is quite large to apply GPBO. 

In \cite{hsic:feature_selection}, the authors introduce the use of HSIC for feature selection and in \cite{hsic}, HSIC based feature selection is used in the context of optimization. The idea is to compute $S_{\rx_i, \mathbf{Y}}$ for each variable involved in the optimization and to discard low $S_{\rx_i, \mathbf{Y}}$ variables from it. More specifically, we fix the discarded variables to an arbitrary value, and then the optimization algorithm is applied only in the dimension of the high $S_{\rx_i, \mathbf{Y}}$ variables. 

This methodology is particularly suited to hyperparameter optimization. In this work, we have emphasized the ability of HSIC to identify the most important hyperparameters. It allows performing relevant HSIC driven hyperparameter selection, which can overcome optimization in too high dimensional hyperparameter spaces. We go further and present a two-step optimization. We optimize the most relevant hyperparameters but also fine-tune less important hyperparameters in a second optimization step. As a result, the problematic optimization in high dimension is split into two easier optimization steps:
\begin{itemize}
    \item[1] Optimization in the reduced yet impactful hyperparameter space, which has reasonable dimension. It allows applying GPBO despite the initially large dimension of the hyperparameter space. At the end of this step, optimal values are selected for the most impactful hyperparameters.
    \item[2] Optimization on the remaining dimensions. In our case, GPBO can be reasonably applied in this space, but note that we might have hyperparameter spaces whose initial dimension is so high that after the first step, the remaining dimensions to optimize could still be too numerous to perform GPBO. In that case, less refined but more robust hyperparameter optimization algorithms (like random search or Tree Parzen Estimators \citep{algoho}) could be applied, which would not be so much of a problem since remaining hyperparameters are less impactful.
\end{itemize}

For the first step, values have to be chosen for less impactful hyperparameters that are not involved in the optimization. In \cite{hsic}, the authors choose the values yielding the best output after the initial random search. Here, the value selection method that aims at improving execution speed, introduced in Section \ref{sect:reduction}, integrates perfectly with this two-step optimization. Following this method brings two advantages. First, we can obtain more cost-effective neural networks if we keep these values through the two optimization step. Second, if we do not care so much about execution speed but only look for accuracy, still fixing these values during the first optimization step improves the training speed and thereby global hyperparameter optimization time. 

The rest of the low $S_{\rx_i, \mathbf{Y}}$ hyperparameters value can be set as those of the hyperparameter configuration yielding the best error. There is one last attention point: one has to be careful about interactions between low  $S_{\rx_i, \mathbf{Y}}$ hyperparameters. If two low HSIC hyperparameters $\rx_i$ and $\rx_j$ are found to interact, like discussed in section \ref{sect:interactions}, and $\rx_i$ has an impact on execution speed, the value of $\rx_j$ must be chosen so that value of the pair $(\rx_i, \rx_j)$ is close to the value of the hyperparameter configuration of a low error neural network. The two-step optimization is summarized in Algorithm \ref{alg:tso}. \\

\begin{algorithm}[h!]
   \caption{ \small Two-step Optimization}
   \label{alg:tso}
    \begin{algorithmic}[1]
   \State {\bfseries Inputs: } hyperparameter search space $\mathcal{H} = \mathcal{X}_1 \times ... \times \mathcal{X}_{n_h}$, $n_s$
   \State Apply Algorithm \ref{alg:hsic}: "Evaluation of HSIC in hyperparameter analysis".
   \State Perform interval reduction (for cost efficiency and stability), as in Sections \ref{sect:reduction} and \ref{sect:reduction_d}.
   \State Select values for less impactful hyperparameters that improve execution speed, taking care of interaction, like discussed in Section \ref{sect:interactions}.\\
   // \texttt{Step 1:}
   \State Apply GPBO to the most impactful hyperparameters.\\
   // \texttt{Step 2:}
   \If{goal = accuracy and execution speed}
       \State Keep the optimal values of step 1 and the values of less impactful hyperparameters that improve execution speed. Apply GPBO to the remaining dimensions. 
   \ElsIf{goal = accuracy only}
        \State Keep the optimal values of step 1. Apply GPBO to the remaining dimensions. 
   \EndIf
    \end{algorithmic}
\end{algorithm}

We evaluate this two-step optimization on our three data sets. For each of these, we consider $4$ baselines. For each of these baselines, we report the \textit{test} error (the metric is accuracy for MNIST and Cifar10 and MSE for Bateman), the number of parameters of the best models, and their FLOPs. 
\begin{itemize}
    \item Random search: The result of the random search of $1000$ configurations plus $200$ additional configurations for a total of $n_s=1200$ points. 
    \item Full GPBO: Gaussian Processes-based Bayesian Optimization, conducted on the full hyperparameter space, without any analysis based on HSIC. We initialize the optimization with $50$ random configurations and perform the optimization for $50$ iterations (enough to reach convergence). 
    \item TS-GPBO (acc): Two-Step GPBO described in Algorithm \ref{alg:tso}, with goal = accuracy. HSIC are estimated using a first random search of $n_s=1000$ points. Steps 1 and 2 are run for $25$ iterations. 
    \item TS-GPBO (acc + speed): Two-Step GPBO described in Algorithm \ref{alg:tso}, with goal = accuracy and execution speed.
\end{itemize}
Random search (ran using 100 parallel jobs for MNIST and Bateman and 24 for Cifar10) took between $2$ and $3$ days depending on the data set, full GPBO between $3$ and $4$ days and TS-GPBO between $3$ and $4$ days as well ($2-3$ days for the initial random search and $1$ day for the two steps of GPBO). Time measure is coarse because not all the training has been conducted on the same architectures (Sandy Bridge CPUs, Nvidia Tesla V100, and Nvidia Tesla P100 GPUs), even within the same baseline, for cluster accessibility reasons. 

We chose the number of total model evaluations for each baseline to obtain approximately the same total execution time. The differences between the number of evaluations, despite identical total execution time, can be explained by different factors. First, the random search can be fully executed in parallel, while GPBO is sequential. Second, step 1 of TS-GPBO always chooses values for non-optimized hyperparameters that improve execution speed and training time. As a result, step 1 is quite fast. Besides, experiments show that step 2 usually converges faster, in terms of the number of evaluations, than full GPBO to the reported minimum, perhaps because the optimal values found during step 1 make step 2 begin close to an optimum. The results of $5$ repetitions (except for random search) of each baseline can be found in Table \ref{tab:results}.

\begin{table}[!ht]
\resizebox{\textwidth}{!}{%
\centering
    \begin{tabular*}{\textwidth}{lllll}
      data set & baseline & test metric & params & MFLOPs \\
      \hline
      MNIST   & RS  & 98.36 & 6,267,103 & 12,709 ($\times 41$)\\
       -    & full GPBO  & \textbf{98.42 }$\pm$ 0.05 & 10,271,367 & 20,534 ($\times 67$)\\
       -   & TS-GPBO (acc + speed)&  \textbf{98.42} $\pm$ 0.02 & \textbf{151,306} & \textbf{307} ($\times 1$)\\
       \hline
      Cifar10   & RS & 81.8 & 99,444,880 & 1,832,615 ($\times 11$)\\
       -    & full GPBO & \textbf{82.73} $\pm$ 1.45 & 71,111,761 & 1,441,230 ($\times 8$)\\
       -   & TS-GPBO (acc) & \textbf{82.60} $\pm$ 0.58 & 9,604,539& 650,269 ($\times 4$) \\
        -   & TS-GPBO (acc + speed) & 79.34 $\pm$ 0.15& \textbf{9,281,258} & \textbf{178,621} ($\times 1$)\\
        \hline
      Bateman   & RS & \textbf{1.99} $\times 10^{-4}$ & 1,259,140 & 2,516 ($\times 359$)\\
       -    & full GPBO  & 2.94 $\pm$ 0.42 $\times 10^{-4}$ & 1,588,215 & 3,173 ($\times 453$)\\
       -   & TS-GPBO (acc + speed) &  3.49 $\pm$ 0.31 $\times 10^{-4}$ & \textbf{3,291} & \textbf{7} ($\times 1$)\\
      \\
    \end{tabular*}}
    
    \caption{Results of hyperparameter optimization for Random Search (RS), Gaussian Processes based Bayesian Optimization on the full hyperparameter space (full GPBO) and Two-Steps Gaussian Processes based Bayesian Optimization (TS-GPBO). The mean $\pm$ standard deviation across $5$ repetitions is displayed for the test metric. For the number of parameters and FLOPs, the maximum value obtained across repetitions is reported because it illustrates the worst scenario that can happen for execution speed, and how much our method prevents it.}
    \label{tab:results}
\end{table}

Results show that except for Cifar10, TS+GPBO yields very competitive neural networks while having far fewer parameters and FLOPs. For MNIST, TS+GPBO model has $\approx$ 66 and 41 times fewer parameters and FLOPs than full GPBO and random search. For Bateman, these factors are 482 and 380. An oversized initial hyperparameter search space could explain such a high factor. Still, a reasonable size for the search space cannot be found \textit{a priori}, and our method makes hyperparameter optimization robust to such bad \textit{a priori} choices. Note that for these cases, we only reported results of TS-GPBO (accuracy + speed) because the results of this baseline were already satisfying, and TS-GPBO (accuracy) did not bring significant improvement. For the particular case of Cifar10, TS-GPBO (accuracy) and (accuracy + speed) both find a model which has 11 and 9 times fewer parameters than random search and full GPBO. TS-GPBO (accuracy) finds a model with $\approx$ 3 and 2 fewer FLOPs than random search and full GPBO while these factors are 10 and 8 for (accuracy + speed). Full GPBO and TS-GPBO (accuracy) achieve comparable accuracy, but the standard deviation for full GPBO is $2.5$ times higher than for TS-GPBO (accuracy), which demonstrates the robustness of TS-GPBO (accuracy). Even if execution time is not an explicitly desired output of TS-GPBO (accuracy), the first step of TS-GPBO, which selects values that improve execution time, seems to bias the optimization towards more cost-effective models, as the final number of parameters and FLOPs shows. All these results have been allowed thanks to information given by HSIC analysis. Hence, TS-GPBO outputs competitive and cost-effective models but also grants a better knowledge of hyperparameters interaction in these machine learning problems, as opposed to random search and full GPBO.

\section{Discussion and Perspectives}

Many techniques have already been introduced to handle hyperparameter optimization, but they often suffer from a lack of interpretability and interactivity. In this work, we tackled these problems by proposing an HSIC based goal-oriented global sensitivity analysis applied to hyperparameter search spaces. We showcased how we can use this information by improving the stability of training instances and the cost efficiency of trained networks. We also introduced an interpretable hyperparameter optimization methodology that yields competitive and cost-effective neural networks based on feature selection.  

\subsection{Impact for scientific machine learning}

These findings are of interest to the machine learning community. Though the presented methodologies can be taken as contributions by themselves, they should also be understood as demonstrations that HSIC based goal-oriented global sensitivity analysis is interesting and valuable for hyperparameter optimization. In the end, an important outcome of this work was to make an insightful tool, HSIC, ready for use in hyperparameter optimization.

This work also impacts scientific computing since it tackles the trade-off between accuracy and cost-efficiency of neural networks. Indeed, we obtained lighter networks without significantly affecting the error, which is the ideal goal for high-performance computing.

\subsection{Other comments}

Other points can be made regarding the presented results and the potential follow-up work. They can be grouped into the following topics: \\

\textbf{Hyperparameters modeling choice.} HSIC is a powerful tool that is widely used for sensitivity analysis as a dependence measure. Its application to hyperparameter optimization required some work, especially regarding the complex structure of hyperparameter space. To achieve this goal, we made some modeling choices, such as applying $\Phi_{\rx_i}$ to map hyperparameter $\rx_i$ to a uniform random variable. The good results obtained in Section \ref{sect:tso} validate not only the usage of information given by HSIC for hyperparameter analysis but also this modeling choice.\\

\textbf{Automating Two-Step Gaussian Process-based Bayesian Optimization.} In this work, we presented methodologies for exploiting HSIC information that involved human intervention. Indeed, someone has to actively decide which hyperparameter deserves to be considered as more or less impactful. Nevertheless, one advantage of HSIC is that it is a scalar metric. One could construct an HSIC based hyperparameter optimization by setting a threshold above which hyperparameters are considered impactful. It would lead to an end-to-end automatic yet interpretable hyperparameter optimization algorithm. Though \cite{hsic} use the idea of a threshold, its application to hyperparameter optimization has not been studied in this paper and could be part of future works.\\

\textbf{Other dependence measures.} In this work, we used HSIC because it is a popular and flexible dependence measure. Our derivations for its application to hyperparameter analysis still hold for any other dependence measure sharing the same properties as HSIC, though studies of different dependence measures is beyond the scope of this paper. \\

\textbf{Hyperparameter optimization speed up.} We presented some ways of using HSIC in hyperparameter optimization, but this paper mainly emphasized the possibility of exploiting it in order to find lighter models. We are aware that execution speed is not always a goal for machine learning practitioners. Still, machine learning practitioners are always concerned about training speed. The first step of TS-GPBO (accuracy) demonstrated the possibility to use HSIC to improve training speed without hurting the final accuracy, so even if final execution speed is not a goal, TS-GPBO made it interesting to use HSIC for that purpose. It would even be possible to go further and to apply parallel GPBO like described in \cite{hsic:spearmint}, or to use Hyperband on the initial random search since HSIC computation only relies on the error of the $p$-\% best neural networks.\\

\textbf{Further execution time improvement.} One advantage of execution time improvement obtained thanks to HSIC is that it only relies on choices for the conception of the neural network. Therefore, additional improvements could be made by applying other techniques like quantization, weights pruning, or multi-objective hyperparameter optimization. 

\section{Conclusion}

Hyperparameter optimization is a very important step of machine learning applications, ordinarily conducted in a black-box fashion. Using an approach based on goal-oriented global sensitivity analysis, we show that we can make hyperparameter optimization more interpretable. In particular, we adapt Hilbert Schmidt Independence Criterion, a statistical dependence measure used in sensitivity analysis, to hyperparameter spaces that can be complex and awkward due to the different nature of hyperparameters (continuous or categorical) and their interactions and inter-dependencies. Its use for hyperparameter analysis is demonstrated on various case studies. In particular, it allows constructing an original and interpretable two-step hyperparameter optimization methodology based on feature selection that improves neural networks' execution speed as well as test error.

\newpage
\bibliography{refs.bib}


\begin{thebibliography}{34}
\ifx \bisbn   \undefined \def \bisbn  #1{ISBN #1}\fi
\ifx \binits  \undefined \def \binits#1{#1}\fi
\ifx \bauthor  \undefined \def \bauthor#1{#1}\fi
\ifx \batitle  \undefined \def \batitle#1{#1}\fi
\ifx \bjtitle  \undefined \def \bjtitle#1{#1}\fi
\ifx \bvolume  \undefined \def \bvolume#1{\textbf{#1}}\fi
\ifx \byear  \undefined \def \byear#1{#1}\fi
\ifx \bissue  \undefined \def \bissue#1{#1}\fi
\ifx \bfpage  \undefined \def \bfpage#1{#1}\fi
\ifx \blpage  \undefined \def \blpage #1{#1}\fi
\ifx \burl  \undefined \def \burl#1{\textsf{#1}}\fi
\ifx \doiurl  \undefined \def \doiurl#1{\url{https://doi.org/#1}}\fi
\ifx \betal  \undefined \def \betal{\textit{et al.}}\fi
\ifx \binstitute  \undefined \def \binstitute#1{#1}\fi
\ifx \binstitutionaled  \undefined \def \binstitutionaled#1{#1}\fi
\ifx \bctitle  \undefined \def \bctitle#1{#1}\fi
\ifx \beditor  \undefined \def \beditor#1{#1}\fi
\ifx \bpublisher  \undefined \def \bpublisher#1{#1}\fi
\ifx \bbtitle  \undefined \def \bbtitle#1{#1}\fi
\ifx \bedition  \undefined \def \bedition#1{#1}\fi
\ifx \bseriesno  \undefined \def \bseriesno#1{#1}\fi
\ifx \blocation  \undefined \def \blocation#1{#1}\fi
\ifx \bsertitle  \undefined \def \bsertitle#1{#1}\fi
\ifx \bsnm \undefined \def \bsnm#1{#1}\fi
\ifx \bsuffix \undefined \def \bsuffix#1{#1}\fi
\ifx \bparticle \undefined \def \bparticle#1{#1}\fi
\ifx \barticle \undefined \def \barticle#1{#1}\fi
\bibcommenthead
\ifx \bconfdate \undefined \def \bconfdate #1{#1}\fi
\ifx \botherref \undefined \def \botherref #1{#1}\fi
\ifx \url \undefined \def \url#1{\textsf{#1}}\fi
\ifx \bchapter \undefined \def \bchapter#1{#1}\fi
\ifx \bbook \undefined \def \bbook#1{#1}\fi
\ifx \bcomment \undefined \def \bcomment#1{#1}\fi
\ifx \oauthor \undefined \def \oauthor#1{#1}\fi
\ifx \citeauthoryear \undefined \def \citeauthoryear#1{#1}\fi
\ifx \endbibitem  \undefined \def \endbibitem {}\fi
\ifx \bconflocation  \undefined \def \bconflocation#1{#1}\fi
\ifx \arxivurl  \undefined \def \arxivurl#1{\textsf{#1}}\fi
\csname PreBibitemsHook\endcsname

\bibitem{deeplearningbook}
\begin{bbook}
\bauthor{\bsnm{Goodfellow}, \binits{I.}},
\bauthor{\bsnm{Bengio}, \binits{Y.}},
\bauthor{\bsnm{Courville}, \binits{A.}}:
\bbtitle{Deep Learning},
(\byear{2016}).
\bcomment{\url{http://www.deeplearningbook.org}}
\end{bbook}
\endbibitem

\bibitem{hsicgretton}
\begin{bchapter}
\bauthor{\bsnm{Gretton}, \binits{A.}},
\bauthor{\bsnm{Bousquet}, \binits{O.}},
\bauthor{\bsnm{Smola}, \binits{A.}},
\bauthor{\bsnm{Sch\"{o}lkopf}, \binits{B.}}:
\bctitle{Measuring statistical dependence with hilbert-schmidt norms}.
In: \bbtitle{Proceedings of the 16th International Conference on Algorithmic
  Learning Theory}.
\bsertitle{ALT’05},
pp. \bfpage{63}--\blpage{77}.
\bpublisher{Springer},
\blocation{Berlin, Heidelberg}
(\byear{2005}).
\doiurl{10.1007/11564089_7}.
\burl{https://doi.org/10.1007/11564089_7}
\end{bchapter}
\endbibitem

\bibitem{two-sample-problem}
\begin{bchapter}
\bauthor{\bsnm{Gretton}, \binits{A.}},
\bauthor{\bsnm{Borgwardt}, \binits{K.}},
\bauthor{\bsnm{Rasch}, \binits{M.}},
\bauthor{\bsnm{Sch\"{o}lkopf}, \binits{B.}},
\bauthor{\bsnm{Smola}, \binits{A.J.}}:
\bctitle{A kernel method for the two-sample-problem}.
In: \beditor{\bsnm{Sch\"{o}lkopf}, \binits{B.}},
\beditor{\bsnm{Platt}, \binits{J.C.}},
\beditor{\bsnm{Hoffman}, \binits{T.}} (eds.)
\bbtitle{Advances in Neural Information Processing Systems 19},
(\byear{2007}).
\burl{http://papers.nips.cc/paper/3110-a-kernel-method-for-the-two-sample-problem.pdf}
\end{bchapter}
\endbibitem

\bibitem{batchnorm}
\begin{bchapter}
\bauthor{\bsnm{Ioffe}, \binits{S.}},
\bauthor{\bsnm{Szegedy}, \binits{C.}}:
\bctitle{Batch normalization: Accelerating deep network training by reducing
  internal covariate shift}.
In: \beditor{\bsnm{Bach}, \binits{F.}},
\beditor{\bsnm{Blei}, \binits{D.}} (eds.)
\bbtitle{Proceedings of the 32nd International Conference on Machine Learning}.
\bsertitle{Proceedings of Machine Learning Research},
vol. \bseriesno{37},
pp. \bfpage{448}--\blpage{456}.
\bpublisher{PMLR},
\blocation{Lille, France}
(\byear{2015}).
\burl{https://proceedings.mlr.press/v37/ioffe15.html}
\end{bchapter}
\endbibitem

\bibitem{adam}
\begin{bchapter}
\bauthor{\bsnm{Kingma}, \binits{D.P.}},
\bauthor{\bsnm{Ba}, \binits{J.}}:
\bctitle{Adam: A method for stochastic optimization}.
In: \bbtitle{ICLR (Poster)}
(\byear{2015}).
\burl{http://arxiv.org/abs/1412.6980}
\end{bchapter}
\endbibitem

\bibitem{efficientnet}
\begin{bchapter}
\bauthor{\bsnm{Tan}, \binits{M.}},
\bauthor{\bsnm{Le}, \binits{Q.}}:
\bctitle{{E}fficient{N}et: Rethinking model scaling for convolutional neural
  networks}.
In: \beditor{\bsnm{Chaudhuri}, \binits{K.}},
\beditor{\bsnm{Salakhutdinov}, \binits{R.}} (eds.)
\bbtitle{Proceedings of the 36th International Conference on Machine Learning}.
\bsertitle{Proceedings of Machine Learning Research},
vol. \bseriesno{97},
pp. \bfpage{6105}--\blpage{6114}.
\bpublisher{PMLR},
\blocation{Long Beach, California, USA}
(\byear{2019}).
\burl{http://proceedings.mlr.press/v97/tan19a.html}
\end{bchapter}
\endbibitem

\bibitem{random-search}
\begin{barticle}
\bauthor{\bsnm{Bergstra}, \binits{J.}},
\bauthor{\bsnm{Bengio}, \binits{Y.}}:
\batitle{Random search for hyper-parameter optimization}.
\bjtitle{Journal of Machine Learning Research}
\bvolume{13}(\bissue{10}),
\bfpage{281}--\blpage{305}
(\byear{2012})
\end{barticle}
\endbibitem

\bibitem{succhalv}
\begin{bchapter}
\bauthor{\bsnm{Jamieson}, \binits{K.}},
\bauthor{\bsnm{Talwalkar}, \binits{A.}}:
\bctitle{Non-stochastic best arm identification and hyperparameter
  optimization}.
In: \beditor{\bsnm{Gretton}, \binits{A.}},
\beditor{\bsnm{Robert}, \binits{C.C.}} (eds.)
\bbtitle{Proceedings of the 19th International Conference on Artificial
  Intelligence and Statistics}.
\bsertitle{Proceedings of Machine Learning Research},
vol. \bseriesno{51},
pp. \bfpage{240}--\blpage{248}.
\bpublisher{PMLR},
\blocation{Cadiz, Spain}
(\byear{2016}).
\burl{http://proceedings.mlr.press/v51/jamieson16.html}
\end{bchapter}
\endbibitem

\bibitem{hyperband}
\begin{barticle}
\bauthor{\bsnm{Li}, \binits{L.}},
\bauthor{\bsnm{Jamieson}, \binits{K.}},
\bauthor{\bsnm{DeSalvo}, \binits{G.}},
\bauthor{\bsnm{Rostamizadeh}, \binits{A.}},
\bauthor{\bsnm{Talwalkar}, \binits{A.}}:
\batitle{Hyperband: A novel bandit-based approach to hyperparameter
  optimization}.
\bjtitle{Journal of Machine Learning Research}
\bvolume{18}(\bissue{185}),
\bfpage{1}--\blpage{52}
(\byear{2018})
\end{barticle}
\endbibitem

\bibitem{bayesian_optim}
\begin{bchapter}
\bauthor{\bsnm{Mockus}, \binits{J.}}:
\bctitle{On bayesian methods for seeking the extremum}.
In: \bbtitle{Proceedings of the IFIP Technical Conference},
pp. \bfpage{400}--\blpage{404}.
\bpublisher{Springer},
\blocation{Berlin, Heidelberg}
(\byear{1974})
\end{bchapter}
\endbibitem

\bibitem{boreview}
\begin{barticle}
\bauthor{\bsnm{Shahriari}, \binits{B.}},
\bauthor{\bsnm{Swersky}, \binits{K.}},
\bauthor{\bsnm{Wang}, \binits{Z.}},
\bauthor{\bsnm{Adams}, \binits{R.P.}},
\bauthor{\bparticle{de} \bsnm{Freitas}, \binits{N.}}:
\batitle{Taking the human out of the loop: A review of bayesian optimization}.
\bjtitle{Proceedings of the IEEE}
\bvolume{104},
\bfpage{148}--\blpage{175}
(\byear{2016})
\end{barticle}
\endbibitem

\bibitem{hsic:spearmint}
\begin{bchapter}
\bauthor{\bsnm{Snoek}, \binits{J.}},
\bauthor{\bsnm{Larochelle}, \binits{H.}},
\bauthor{\bsnm{Adams}, \binits{R.P.}}:
\bctitle{Practical bayesian optimization of machine learning algorithms}.
In: \bbtitle{Proceedings of the 25th International Conference on Neural
  Information Processing Systems - Volume 2}.
\bsertitle{NIPS'12},
pp. \bfpage{2951}--\blpage{2959}.
\bpublisher{Curran Associates Inc.},
\blocation{Red Hook, NY, USA}
(\byear{2012})
\end{bchapter}
\endbibitem

\bibitem{algoho}
\begin{bchapter}
\bauthor{\bsnm{Bergstra}, \binits{J.S.}},
\bauthor{\bsnm{Bardenet}, \binits{R.}},
\bauthor{\bsnm{Bengio}, \binits{Y.}},
\bauthor{\bsnm{K\'{e}gl}, \binits{B.}}:
\bctitle{Algorithms for hyper-parameter optimization}.
In: \beditor{\bsnm{Shawe-Taylor}, \binits{J.}},
\beditor{\bsnm{Zemel}, \binits{R.S.}},
\beditor{\bsnm{Bartlett}, \binits{P.L.}},
\beditor{\bsnm{Pereira}, \binits{F.}},
\beditor{\bsnm{Weinberger}, \binits{K.Q.}} (eds.)
\bbtitle{Advances in Neural Information Processing Systems 24},
(\byear{2011}).
\burl{http://papers.nips.cc/paper/4443-algorithms-for-hyper-parameter-optimization.pdf}
\end{bchapter}
\endbibitem

\bibitem{dnnho}
\begin{bchapter}
\bauthor{\bsnm{Snoek}, \binits{J.}},
\bauthor{\bsnm{Rippel}, \binits{O.}},
\bauthor{\bsnm{Swersky}, \binits{K.}},
\bauthor{\bsnm{Kiros}, \binits{R.}},
\bauthor{\bsnm{Satish}, \binits{N.}},
\bauthor{\bsnm{Sundaram}, \binits{N.}},
\bauthor{\bsnm{Patwary}, \binits{M.}},
\bauthor{\bsnm{Prabhat}, \binits{M.}},
\bauthor{\bsnm{Adams}, \binits{R.}}:
\bctitle{Scalable bayesian optimization using deep neural networks}.
In: \beditor{\bsnm{Bach}, \binits{F.}},
\beditor{\bsnm{Blei}, \binits{D.}} (eds.)
\bbtitle{Proceedings of the 32nd International Conference on Machine Learning}.
\bsertitle{Proceedings of Machine Learning Research},
vol. \bseriesno{37},
pp. \bfpage{2171}--\blpage{2180}.
\bpublisher{PMLR},
\blocation{Lille, France}
(\byear{2015}).
\burl{http://proceedings.mlr.press/v37/snoek15.html}
\end{bchapter}
\endbibitem

\bibitem{hsic:depthwise}
\begin{botherref}
\oauthor{\bsnm{Chollet}, \binits{F.}}:
Xception: Deep learning with depthwise separable convolutions.
CoRR
\textbf{abs/1610.02357}
(2016)
{\href{https://arxiv.org/abs/1610.02357}{{arXiv:1610.02357}}}
\end{botherref}
\endbibitem

\bibitem{hsic:evo-nas}
\begin{barticle}
\bauthor{\bsnm{Stanley}, \binits{K.O.}},
\bauthor{\bsnm{Miikkulainen}, \binits{R.}}:
\batitle{Evolving neural networks through augmenting topologies}.
\bjtitle{Evol. Comput.}
\bvolume{10}(\bissue{2}),
\bfpage{99}--\blpage{127}
(\byear{2002}).
\doiurl{10.1162/106365602320169811}
\end{barticle}
\endbibitem

\bibitem{hsic:nasbot}
\begin{bchapter}
\bauthor{\bsnm{Kandasamy}, \binits{K.}},
\bauthor{\bsnm{Neiswanger}, \binits{W.}},
\bauthor{\bsnm{Schneider}, \binits{J.}},
\bauthor{\bsnm{P\'{o}czos}, \binits{B.}},
\bauthor{\bsnm{Xing}, \binits{E.P.}}:
\bctitle{Neural architecture search with bayesian optimisation and optimal
  transport}.
In: \bbtitle{Proceedings of the 32nd International Conference on Neural
  Information Processing Systems}.
\bsertitle{NIPS'18},
pp. \bfpage{2020}--\blpage{2029}.
\bpublisher{Curran Associates Inc.},
\blocation{Red Hook, NY, USA}
(\byear{2018})
\end{bchapter}
\endbibitem

\bibitem{hsic:reinf}
\begin{bchapter}
\bauthor{\bsnm{Pham}, \binits{H.}},
\bauthor{\bsnm{Guan}, \binits{M.}},
\bauthor{\bsnm{Zoph}, \binits{B.}},
\bauthor{\bsnm{Le}, \binits{Q.}},
\bauthor{\bsnm{Dean}, \binits{J.}}:
\bctitle{Efficient neural architecture search via parameters sharing}.
\bsertitle{Proceedings of Machine Learning Research},
vol. \bseriesno{80},
pp. \bfpage{4095}--\blpage{4104}.
\bpublisher{PMLR},
\blocation{Stockholmsmässan, Stockholm Sweden}
(\byear{2018}).
\burl{http://proceedings.mlr.press/v80/pham18a.html}
\end{bchapter}
\endbibitem

\bibitem{hsic:mnasnet}
\begin{botherref}
\oauthor{\bsnm{Tan}, \binits{M.}},
\oauthor{\bsnm{Chen}, \binits{B.}},
\oauthor{\bsnm{Pang}, \binits{R.}},
\oauthor{\bsnm{Vasudevan}, \binits{V.}},
\oauthor{\bsnm{Le}, \binits{Q.V.}}:
Mnasnet: Platform-aware neural architecture search for mobile.
CoRR
\textbf{abs/1807.11626}
(2018)
{\href{https://arxiv.org/abs/1807.11626}{{arXiv:1807.11626}}}
\end{botherref}
\endbibitem

\bibitem{hsic:nas-survey}
\begin{barticle}
\bauthor{\bsnm{Elsken}, \binits{T.}},
\bauthor{\bsnm{Metzen}, \binits{J.H.}},
\bauthor{\bsnm{Hutter}, \binits{F.}}:
\batitle{Neural architecture search: A survey}.
\bjtitle{Journal of Machine Learning Research}
\bvolume{20}(\bissue{55}),
\bfpage{1}--\blpage{21}
(\byear{2019})
\end{barticle}
\endbibitem

\bibitem{hsic:futureofsas}
\begin{barticle}
\bauthor{\bsnm{Razavi}, \binits{S.}},
\bauthor{\bsnm{Jakeman}, \binits{A.}},
\bauthor{\bsnm{Saltelli}, \binits{A.}},
\bauthor{\bsnm{Prieur}, \binits{C.}},
\bauthor{\bsnm{Iooss}, \binits{B.}},
\bauthor{\bsnm{Borgonovo}, \binits{E.}},
\bauthor{\bsnm{Plischke}, \binits{E.}},
\bauthor{\bsnm{{Lo Piano}}, \binits{S.}},
\bauthor{\bsnm{Iwanaga}, \binits{T.}},
\bauthor{\bsnm{Becker}, \binits{W.}},
\bauthor{\bsnm{Tarantola}, \binits{S.}},
\bauthor{\bsnm{Guillaume}, \binits{J.H.A.}},
\bauthor{\bsnm{Jakeman}, \binits{J.}},
\bauthor{\bsnm{Gupta}, \binits{H.}},
\bauthor{\bsnm{Melillo}, \binits{N.}},
\bauthor{\bsnm{Rabitti}, \binits{G.}},
\bauthor{\bsnm{Chabridon}, \binits{V.}},
\bauthor{\bsnm{Duan}, \binits{Q.}},
\bauthor{\bsnm{Sun}, \binits{X.}},
\bauthor{\bsnm{Smith}, \binits{S.}},
\bauthor{\bsnm{Sheikholeslami}, \binits{R.}},
\bauthor{\bsnm{Hosseini}, \binits{N.}},
\bauthor{\bsnm{Asadzadeh}, \binits{M.}},
\bauthor{\bsnm{Puy}, \binits{A.}},
\bauthor{\bsnm{Kucherenko}, \binits{S.}},
\bauthor{\bsnm{Maier}, \binits{H.R.}}:
\batitle{The future of sensitivity analysis: An essential discipline for
  systems modeling and policy support}.
\bjtitle{Environmental Modelling \& Software}
\bvolume{137},
\bfpage{104954}
(\byear{2021}).
\doiurl{10.1016/j.envsoft.2020.104954}
\end{barticle}
\endbibitem

\bibitem{Sobol}
\begin{botherref}
\oauthor{\bsnm{Sobol}, \binits{I.M.}}:
Sensitivity estimates for nonlinear mathematical models.
MMCE
(1),
407--414
(1993)
\end{botherref}
\endbibitem

\bibitem{contrast}
\begin{barticle}
\bauthor{\bsnm{Fort}, \binits{J.-C.}},
\bauthor{\bsnm{Klein}, \binits{T.}},
\bauthor{\bsnm{Rachdi}, \binits{N.}}:
\batitle{New sensitivity analysis subordinated to a contrast}.
\bjtitle{Communications in Statistics - Theory and Methods}
\bvolume{45}(\bissue{15}),
\bfpage{4349}--\blpage{4364}
(\byear{2016})
{\href{https://arxiv.org/abs/https://doi.org/10.1080/03610926.2014.901369}{{https://doi.org/10.1080/03610926.2014.901369}}}.
\doiurl{10.1080/03610926.2014.901369}
\end{barticle}
\endbibitem

\bibitem{borgonovo}
\begin{barticle}
\bauthor{\bsnm{Borgonovo}, \binits{E.}}:
\batitle{A new uncertainty importance measure}.
\bjtitle{Reliability Engineering \& System Safety}
\bvolume{92}(\bissue{6}),
\bfpage{771}--\blpage{784}
(\byear{2007}).
\doiurl{10.1016/j.ress.2006.04.015}
\end{barticle}
\endbibitem

\bibitem{hsic:saltelli-doe}
\begin{barticle}
\bauthor{\bsnm{Saltelli}, \binits{A.}}:
\batitle{Making best use of model evaluations to compute sensitivity indices}.
\bjtitle{Computer Physics Communications}
\bvolume{145}(\bissue{2}),
\bfpage{280}--\blpage{297}
(\byear{2002}).
\doiurl{10.1016/S0010-4655(02)00280-1}
\end{barticle}
\endbibitem

\bibitem{gsahsic}
\begin{botherref}
\oauthor{\bsnm{Da~Veiga}, \binits{S.}}:
Global sensitivity analysis with dependence measures.
Journal of Statistical Computation and Simulation
\textbf{85}
(2013).
\doiurl{10.1080/00949655.2014.945932}
\end{botherref}
\endbibitem

\bibitem{fdivergence}
\begin{barticle}
\bauthor{\bsnm{Csizar}, \binits{I.}}:
\batitle{Information-type measures of difference of probability distributions
  and indirect observation}.
\bjtitle{Studia Scientiarum Mathematicarum Hungarica}
\bvolume{2},
\bfpage{229}--\blpage{318}
(\byear{1967})
\end{barticle}
\endbibitem

\bibitem{IPM}
\begin{barticle}
\bauthor{\bsnm{Müller}, \binits{A.}}:
\batitle{Integral probability metrics and their generating classes of
  functions}.
\bjtitle{Advances in Applied Probability}
\bvolume{29}(\bissue{2}),
\bfpage{429}--\blpage{443}
(\byear{1997}).
\doiurl{10.2307/1428011}
\end{barticle}
\endbibitem

\bibitem{MMD-generalized}
\begin{bchapter}
\bauthor{\bsnm{Fukumizu}, \binits{K.}},
\bauthor{\bsnm{Gretton}, \binits{A.}},
\bauthor{\bsnm{Lanckriet}, \binits{G.R.}},
\bauthor{\bsnm{Sch\"{o}lkopf}, \binits{B.}},
\bauthor{\bsnm{Sriperumbudur}, \binits{B.K.}}:
\bctitle{Kernel choice and classifiability for rkhs embeddings of probability
  distributions}.
In: \beditor{\bsnm{Bengio}, \binits{Y.}},
\beditor{\bsnm{Schuurmans}, \binits{D.}},
\beditor{\bsnm{Lafferty}, \binits{J.D.}},
\beditor{\bsnm{Williams}, \binits{C.K.I.}},
\beditor{\bsnm{Culotta}, \binits{A.}} (eds.)
\bbtitle{Advances in Neural Information Processing Systems 22},
(\byear{2009}).
\burl{http://papers.nips.cc/paper/3750-kernel-choice-and-classifiability-for-rkhs-embeddings-of-probability-distributions.pdf}
\end{bchapter}
\endbibitem

\bibitem{hsic}
\begin{barticle}
\bauthor{\bsnm{Spagnol}, \binits{A.}},
\bauthor{\bsnm{Riche}, \binits{R.L.}},
\bauthor{\bsnm{{Da Veiga}}, \binits{S.}}:
\batitle{Global sensitivity analysis for optimization with variable selection}.
\bjtitle{SIAM/ASA J. Uncertain. Quantification}
\bvolume{7},
\bfpage{417}--\blpage{443}
(\byear{2018})
\end{barticle}
\endbibitem

\bibitem{categ}
\begin{barticle}
\bauthor{\bsnm{Kruskal}, \binits{J.B.}}:
\batitle{Multidimensional scaling by optimizing goodness of fit to a nonmetric
  hypothesis}.
\bjtitle{Psychometrika}
\bvolume{29}(\bissue{1}),
\bfpage{1}--\blpage{27}
(\byear{1964}).
\doiurl{10.1007/BF02289565}.
Accessed 2021-10-12
\end{barticle}
\endbibitem

\bibitem{MCcdf}
\begin{barticle}
\bauthor{\bsnm{Gillespie}, \binits{D.T.}}:
\batitle{A general method for numerically simulating the stochastic time
  evolution of coupled chemical reactions}.
\bjtitle{Journal of Computational Physics}
\bvolume{22}(\bissue{4}),
\bfpage{403}--\blpage{434}
(\byear{1976}).
\doiurl{10.1016/0021-9991(76)90041-3}
\end{barticle}
\endbibitem

\bibitem{hsic:bohb}
\begin{bchapter}
\bauthor{\bsnm{Falkner}, \binits{S.}},
\bauthor{\bsnm{Klein}, \binits{A.}},
\bauthor{\bsnm{Hutter}, \binits{F.}}:
\bctitle{{BOHB}: Robust and efficient hyperparameter optimization at scale}.
\bsertitle{Proceedings of Machine Learning Research},
vol. \bseriesno{80},
pp. \bfpage{1437}--\blpage{1446}.
\bpublisher{PMLR},
\blocation{Stockholmsmässan, Stockholm Sweden}
(\byear{2018}).
\burl{http://proceedings.mlr.press/v80/falkner18a.html}
\end{bchapter}
\endbibitem

\bibitem{hsic:feature_selection}
\begin{bchapter}
\bauthor{\bsnm{Song}, \binits{L.}},
\bauthor{\bsnm{Smola}, \binits{A.}},
\bauthor{\bsnm{Gretton}, \binits{A.}},
\bauthor{\bsnm{Borgwardt}, \binits{K.M.}},
\bauthor{\bsnm{Bedo}, \binits{J.}}:
\bctitle{Supervised feature selection via dependence estimation}.
In: \bbtitle{Proceedings of the 24th International Conference on Machine
  Learning}.
\bsertitle{ICML ’07},
pp. \bfpage{823}--\blpage{830}.
\bpublisher{Association for Computing Machinery},
\blocation{New York, NY, USA}
(\byear{2007}).
\doiurl{10.1145/1273496.1273600}.
\burl{https://doi.org/10.1145/1273496.1273600}
\end{bchapter}
\endbibitem

\end{thebibliography}

\newpage
\section*{Statements \& Declarations}

\paragraph{Ethics approval and Consent to participate :} All authors approve the Committee on Publication Ethics guidelines. They consent to participate.

\paragraph{Consent for publication :}  All authors give explicit consent to submit and they obtained consent from the responsible authorities at the institutes where the work has been carried out, the CEA, Inria and the Ecole Polytechnique.

\paragraph{Availability of data and materials :} The source code is available at \url{https://github.com/paulnovello/goal-oriented-ho}.

\paragraph{Competing interests :} All authors certify that they have no affiliations with or involvement in any organization or entity with any financial interest or non-financial interest in the subject matter or materials discussed in this manuscript.

\paragraph{Fundings :} The research has been supported by the institutes where the authors are affiliated : the CEA, Inria and the Ecole Polytechnique.\\

\paragraph{Authors' contributions :} 

All authors whose names appear on the submission

\begin{itemize}
    \item made substantial contributions to the conception or design of the work; or the acquisition, analysis, or interpretation of data; or the creation of new software used in the work
    \item drafted the work or revised it critically for important intellectual content
    \item approved the version to be published
    \item agree to be accountable for all aspects of the work in ensuring that questions related to the accuracy or integrity of any part of the work are appropriately investigated and resolved.
\end{itemize}

\newpage

\begin{appendices}
\section{Hyperparameters spaces}\label{appA}

In this section, we describe hyperparameters spaces used for each problem in this chapter. Note that hyperparameter \texttt{n\_seeds} denotes the number of random repetitions of the training for each hyperparameter configuration. If a conditional hyperparameter $X_j$ is only involved for some specific values of a main hyperparameter $X_i$, it is displayed with an indent on tab lines below that of $X_i$, with the value of $X_i$ required for $X_j$ to be involved in the training. 

\subsection{Runge and MNIST}

For Runge and MNIST, only fully connected Neural Networks are trained, and the width (\texttt{n\_units}) is the same for every layer.
\begin{table}[h]
  \centering
  \begin{tabular}{llll}
    \toprule
    hyperparameter & type & values for Runge & values for MNIST\\
    \bottomrule
    \texttt{n\_layers} & integer & $\in \{1,...,10\}$ & same \\
    \texttt{n\_units} & integer & $\in \{7,...,512\}$ & $\in \{128,...,1500\}$\\
    \texttt{activation} & categorical & \texttt{elu}, \texttt{relu}, \texttt{tanh} or \texttt{sigmoid} &  same\\
    \texttt{dropout} & boolean & \texttt{true} or \texttt{false} & same\\
    \texttt{    yes:dropout\_rate} & continuous & $\in [0,1]$ & same\\
    \texttt{batch\_norm} & boolean & \texttt{true} or \texttt{false}  & same\\
    \texttt{weights\_reg\_l1} & continuous& $\in [1\times10^{-6},0.1]$ & same\\
    \texttt{weights\_reg\_l2} &continuous & $\in [1\times10^{-6},0.1]$ & same\\
    \texttt{bias\_reg\_l1} &continuous & $\in [1\times10^{-6},0.1]$ & same\\
    \texttt{bias\_reg\_l2} &continuous & $\in [1\times10^{-6},0.1]$ & same\\
    \texttt{batch\_size} & integer&$\in \{1,...,11\}$ & $\in \{1,...,256\}$ \\
    \texttt{loss\_function} & categorical & $L_2$ error or $L_1$ error &  $L_2$ error or crossentropy\\
    \texttt{optimizer} & categorical  & \texttt{adam}, \texttt{sgd}, \texttt{rmsprop} or \texttt{adagrad} & same\\
    \texttt{n\_seeds} & integer &$\in \{1,...,40\}$ & $\in \{1,...,10\}$\\
  \end{tabular}
  \caption{ Hyperparameters values for Runge \& MNIST}
\end{table}

\textbf{Conditional groups:} (see \textbf{(iii)} of Section \ref{sect:cond}) $\mathcal{G}_0$ and $\mathcal{G}_{\texttt{dropout\_rate}}$

\clearpage
\subsection{Bateman}

For Bateman, only fully connected Neural Networks are trained, and the width (\texttt{n\_units}) is the same for every layer.

\begin{table}[!ht]
  \centering
  \begin{tabular}{lll}
    \toprule
    hyperparameter & type & values for Bateman  \\
    \bottomrule
    \texttt{n\_layers} & integer & $\in \{1,...,10\}$ \\
    \texttt{n\_units} & integer & $\in \{7,...,512\}$\\
    \texttt{activation} & categorical & \texttt{elu}, \texttt{relu}, \texttt{tanh} or \texttt{sigmoid}\\
    \texttt{dropout} & boolean & \texttt{true} or \texttt{false} \\
    \texttt{    yes:dropout\_rate} & continuous & $\in [0,1]$ \\
    \texttt{batch\_norm} & boolean & \texttt{true} or \texttt{false} \\
    \texttt{learning\_rate} & continuous& $\in [1\times10^{-6},1\times10^{-2}]$ \\
    \texttt{weights\_reg\_l1} & continuous& $\in [1\times10^{-6},0.1]$ \\
    \texttt{weights\_reg\_l2} &continuous & $\in [1\times10^{-6},0.1]$ \\
    \texttt{bias\_reg\_l1} &continuous & $\in [1\times10^{-6},0.1]$ \\
    \texttt{bias\_reg\_l2} &continuous & $\in [1\times10^{-6},0.1]$ \\
    \texttt{batch\_size} & integer & $\in \{1,...,500\}$ \\
    \texttt{loss\_function} & categorical & $L_2$ error or $L_1$ error  \\
    \texttt{optimizer} & categorical  & \texttt{adam}, \texttt{sgd}, \texttt{rmsprop}, \texttt{adagrad} or \texttt{nadam} \\
    \texttt{    adam:amsgrad} & boolean & \texttt{true} or \texttt{false}  \\
    \texttt{    adam, nadam:1st\_moment\_decay} & continuous & $\in [0.8,1]$ \\
    \texttt{    adam, nadam:2nd\_moment\_decay} & continuous & $\in [0.8,1]$ \\
    \texttt{    rmsprop:centered} & boolean & \texttt{true} or \texttt{false}\\
    \texttt{    sgd:nesterov} & boolean & \texttt{true} or \texttt{false} \\
    \texttt{    sgd, rmsprop:momentum} & continuous & $\in [0.5,0.99]$ \\
    \texttt{n\_seeds} & integer & $\in \{1,...,10\}$\\
  \end{tabular}
  \caption{ Hyperparameters values for Bateman}
\end{table}

\textbf{Conditional groups:} (see \textbf{(iii)} of Section \ref{sect:cond}) $\mathcal{G}_0$, $\mathcal{G}_{\texttt{dropout\_rate}}$, $\mathcal{G}_{\texttt{amsgrad}}$, $\mathcal{G}_{\texttt{centered}}$, $\mathcal{G}_{\texttt{nesterov}}$, $\mathcal{G}_{\texttt{momentum}}$ and \\
$\mathcal{G}_{(\texttt{1st\_moment}, \texttt{2nd\_moment})}$

\clearpage
\subsection{Cifar10}

For Cifar10, we use Convolutional Neural Networks, whose width increases with the depth according to hyperparameters \texttt{stages} and \texttt{stage\_mult}. The first layer has width \texttt{n\_filters}, and then, $\texttt{stages} - 1$ times, the network is widen by a factor \texttt{stage\_mult}. For instance, a neural network with $\texttt{n\_filters} = 20$, $\texttt{n\_layers}=3$, $\texttt{stages}=3$ and $\texttt{stage\_mult}=2$ will have a first layer with $20$ filters, a second layer with $\texttt{n\_filters} \times {stage\_mult} = 40$ filters, and a third layer with $\texttt{n\_filters} \times \texttt{stage\_mult}^{\texttt{stages}-1} = 60$ filters.

\begin{table}[h]
  \centering
  \begin{tabular}{lll}
    \toprule
    hyperparameter & type & values for Cifar10  \\
    \bottomrule
    \texttt{n\_layers} & integer & $\in \{3,...,12\}$ \\
    \texttt{n\_filters} & integer & $\in \{16,...,100\}$\\
    \texttt{stages} & integer & $\in \{1,4\}$\\
    \texttt{stage\_mult} & continuous & $\in [1,3]$\\
    \texttt{kernel\_size} & integer & $\in \{1,5\}$\\
    \texttt{pool\_size} & integer & $\in \{2,5\}$\\
    \texttt{pool\_type} & categorical & \texttt{max} or \texttt{average}\\
    \texttt{activation} & categorical & \texttt{elu}, \texttt{relu}, \texttt{tanh} or \texttt{sigmoid}\\
    \texttt{dropout} & boolean & \texttt{true} or \texttt{false} \\
    \texttt{    yes:dropout\_rate} & continuous & $\in [0,1]$ \\
    \texttt{batch\_norm} & boolean & \texttt{true} or \texttt{false} \\
    \texttt{learning\_rate} & continuous& $\in [1\times10^{-6},1\times10^{-2}]$ \\
    \texttt{weights\_reg\_l1} & continuous& $\in [1\times10^{-6},0.1]$ \\
    \texttt{weights\_reg\_l2} &continuous & $\in [1\times10^{-6},0.1]$ \\
    \texttt{bias\_reg\_l1} &continuous & $\in [1\times10^{-6},0.1]$ \\
    \texttt{bias\_reg\_l2} &continuous & $\in [1\times10^{-6},0.1]$ \\
    \texttt{batch\_size} & integer & $\in \{10,...,128\}$ \\
    \texttt{loss\_function} & categorical & $L_2$ error or crossentropy \\
    \texttt{optimizer} & categorical  & \texttt{adam}, \texttt{sgd}, \texttt{rmsprop}, \texttt{adagrad} or \texttt{nadam} \\
    \texttt{    adam:amsgrad} & boolean & \texttt{true} or \texttt{false}  \\
    \texttt{    adam, nadam:1st\_moment\_decay} & continuous & $\in [0.8,1]$ \\
    \texttt{    adam, nadam:2nd\_moment\_decay} & continuous & $\in [0.8,1]$ \\
    \texttt{    rmsprop:centered} & boolean & \texttt{true} or \texttt{false}\\
    \texttt{    sgd:nesterov} & boolean & \texttt{true} or \texttt{false} \\
    \texttt{    sgd, rmsprop:momentum} & continuous & $\in [0.5,0.99]$ \\
    \texttt{n\_seeds} & integer & $\in \{1,...,10\}$\\
  \end{tabular}
  \caption{ Hyperparameters values for Cifar10}
\end{table}

\textbf{Conditional groups:} (see \textbf{(iii)} of Section \ref{sect:cond}) $\mathcal{G}_0$, $\mathcal{G}_{\texttt{dropout\_rate}}$, $\mathcal{G}_{\texttt{amsgrad}}$, $\mathcal{G}_{\texttt{centered}}$, $\mathcal{G}_{\texttt{nesterov}}$, $\mathcal{G}_{\texttt{momentum}}$ and \\
$\mathcal{G}_{(\texttt{1st\_moment}, \texttt{2nd\_moment})}$

\newpage
\section{HSICs for conditional hyperparameters}\label{appB}

\subsection{MNIST}

For MNIST, there is only one conditional hyperparameter, \texttt{dropout\_rate}, so only one conditional group to consider in order to assess the importance of conditional hyperparameters.

\begin{figure}[!ht]
\centering
     \includegraphics[width=0.4\linewidth]{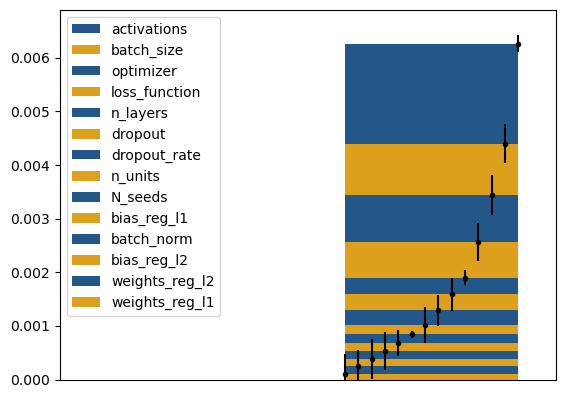}

   \caption{HSICs for $\mathcal{G}_{\texttt{dropout\_rate}}$ of MNIST hyperparameter analysis. Conditional hyperparameter \texttt{dropout\_rate} is not impactful.}
\end{figure}

\newpage
\subsection{Bateman}

\begin{figure}[!ht]
\centering
   \begin{subfigure}{0.32\textwidth}
     \includegraphics[width=1.0\linewidth]{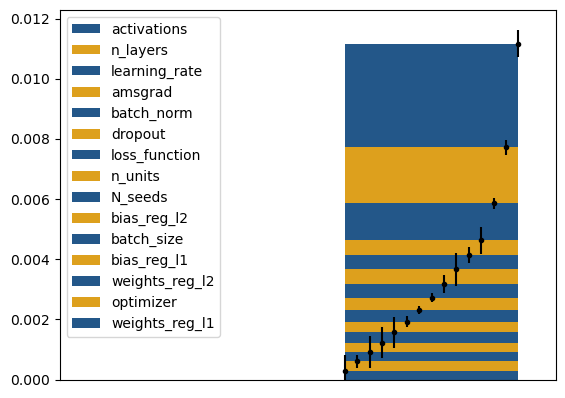}
     \caption{$\mathcal{G}_{\texttt{amsgrad}}$. }
   \end{subfigure}
   \begin{subfigure}{0.32\textwidth}
     \includegraphics[width=1.0\linewidth]{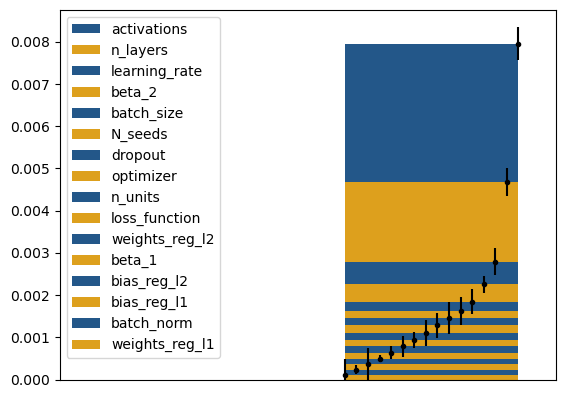}
     \caption{$\mathcal{G}_{(\texttt{1st\_moment}, \texttt{2nd\_moment})}$}
   \end{subfigure}
      \begin{subfigure}{0.32\textwidth}
     \includegraphics[width=1.0\linewidth]{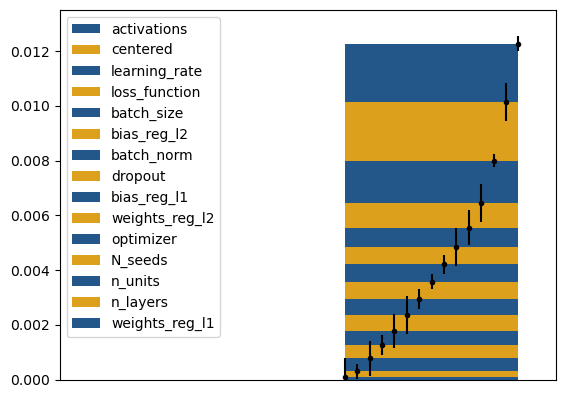}
     \caption{$\mathcal{G}_{\texttt{centered}}$}
   \end{subfigure}
      \begin{subfigure}{0.32\textwidth}
     \includegraphics[width=1.0\linewidth]{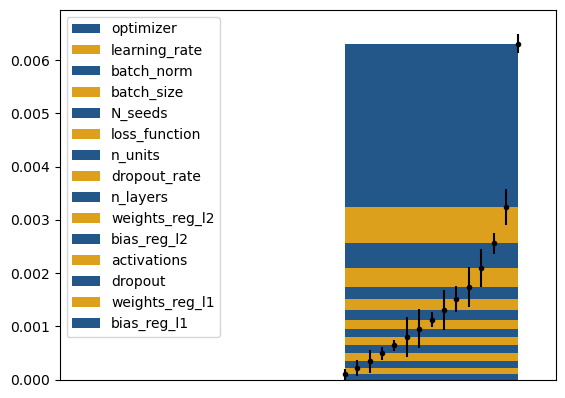}
     \caption{$\mathcal{G}_{\texttt{dropout\_rate}}$}
   \end{subfigure}
  \begin{subfigure}{0.32\textwidth}
     \includegraphics[width=1.0\linewidth]{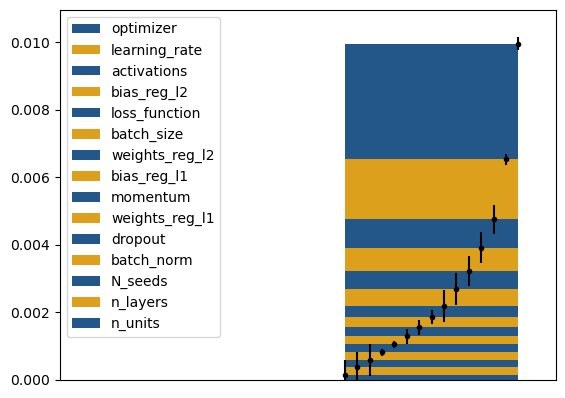}
     \caption{$\mathcal{G}_{\texttt{momentum}}$}
   \end{subfigure}
    \begin{subfigure}{0.32\textwidth}
     \includegraphics[width=1.0\linewidth]{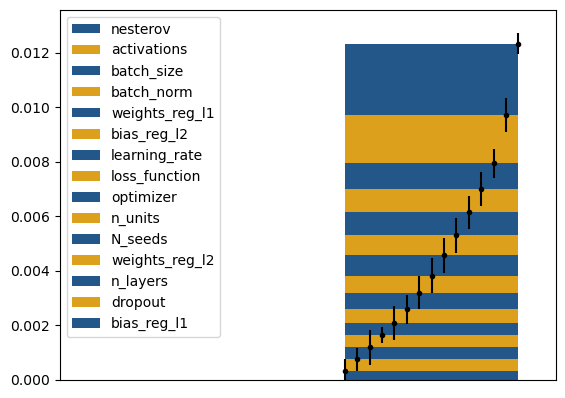}
     \caption{$\mathcal{G}_{\texttt{nesterov}}$}
   \end{subfigure}
   \caption{HSICs for conditional groups of Bateman hyperparameter analysis. (a): \texttt{amsgrad} is not impactful (it is in the estimation noise), (b): \texttt{1st\_moment} is not impactful but \texttt{2nd\_moment} is the fourth most impactful hyperparameter of this group, (c): \texttt{centered} is the second most impactful hyperparameter of this group, (d): \texttt{dropout\_rate} is not impactful, (e): \texttt{momentum} is not impactful, (f): \texttt{nesterov} is the most impactful hyperparameter of this group.}
   \label{fig:appCB}
\end{figure}

For Bateman, there are seven conditional hyperparameter, \texttt{amsgrad}, \texttt{1st\_moment} (\texttt{beta\_1}), \texttt{2nd\_moment} (\texttt{beta\_2}), \texttt{dropout\_rate}, \texttt{centered}, \texttt{momentum}, and \texttt{nesterov}. Six conditional groups, specified in Figure \ref{fig:appCB}, have to be considered in order to assess their importance.

\newpage
\subsection{Cifar10}

\begin{figure}[!ht]
\centering
   \begin{subfigure}{0.32\textwidth}
     \includegraphics[width=1.0\linewidth]{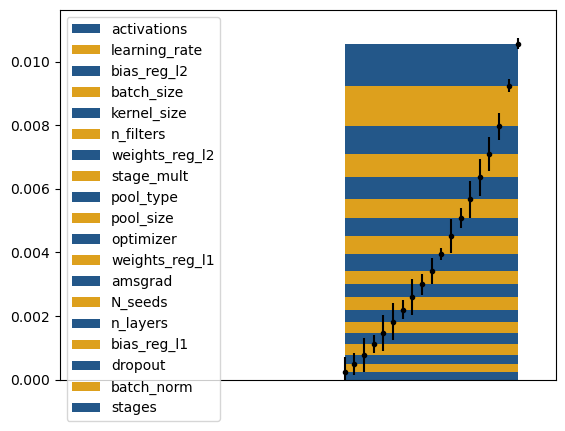}
     \caption{$\mathcal{G}_{\texttt{amsgrad}}$}
   \end{subfigure}
   \begin{subfigure}{0.32\textwidth}
     \includegraphics[width=1.0\linewidth]{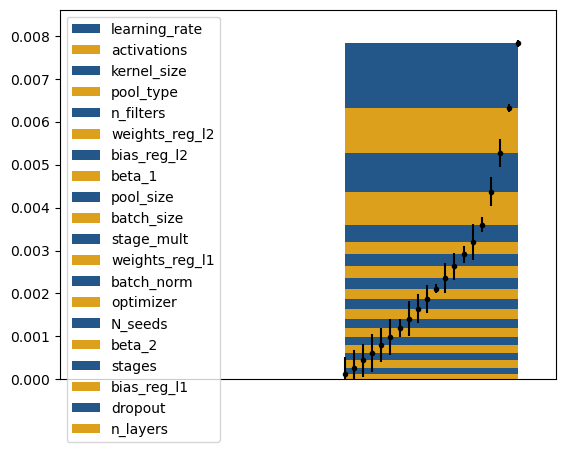}
     \caption{$\mathcal{G}_{(\texttt{1st\_moment}, \texttt{2nd\_moment})}$}
   \end{subfigure}
      \begin{subfigure}{0.32\textwidth}
     \includegraphics[width=1.0\linewidth]{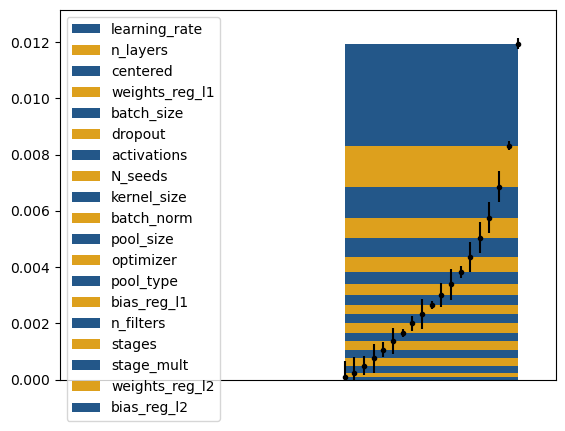}
     \caption{$\mathcal{G}_{\texttt{centered}}$}
   \end{subfigure}
      \begin{subfigure}{0.32\textwidth}
     \includegraphics[width=1.0\linewidth]{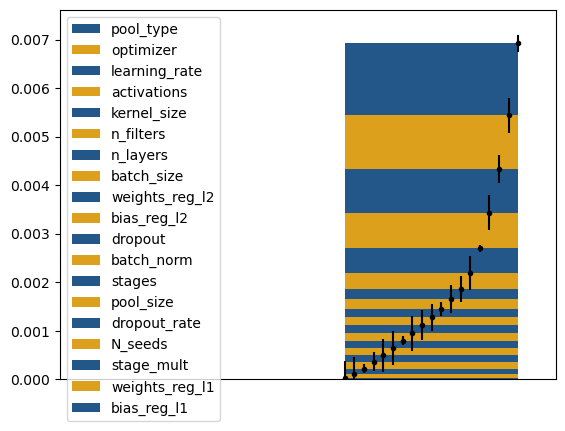}
     \caption{$\mathcal{G}_{\texttt{dropout\_rate}}$}
   \end{subfigure}
  \begin{subfigure}{0.32\textwidth}
     \includegraphics[width=1.0\linewidth]{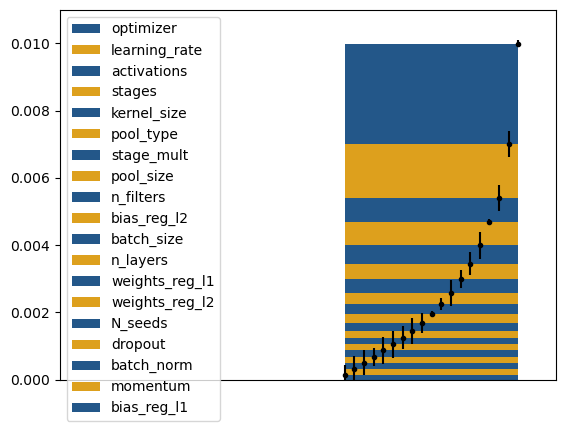}
     \caption{$\mathcal{G}_{\texttt{momentum}}$}
   \end{subfigure}
    \begin{subfigure}{0.32\textwidth}
     \includegraphics[width=1.0\linewidth]{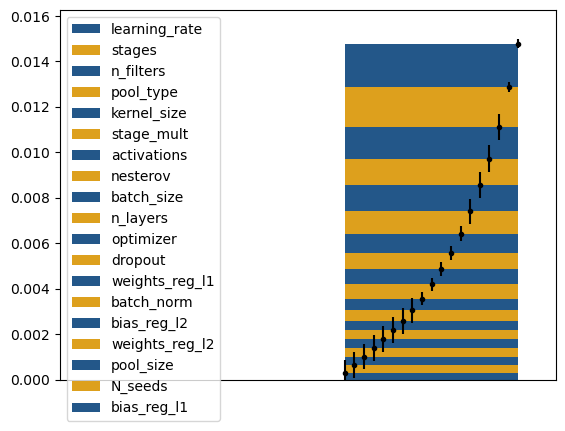}
     \caption{$\mathcal{G}_{\texttt{nesterov}}$}
   \end{subfigure}
   \caption{HSICs for conditional groups of cifar10 hyperparameter analysis. (a): \texttt{amsgrad} is not impactful, (b): \texttt{1st\_moment}, \texttt{2nd\_moment} are not impactful, (c): \texttt{centered} is the third most impactful hyperparameter of this group, (d): \texttt{dropout\_rate} is not impactful, (e): \texttt{momentum} is not impactful, (f): \texttt{nesterov} is not impactful.}
   \label{fig:appCC}
\end{figure}

For Cifar10, there are seven conditional hyperparameter, \texttt{amsgrad}, \texttt{1st\_moment} (\texttt{beta\_1}), \texttt{2nd\_moment} (\texttt{beta\_2}), \texttt{dropout\_rate}, \texttt{centered}, \texttt{momentum}, and \texttt{nesterov}. Six conditional groups, specified in Figure \ref{fig:appCC}, have to be considered in order to assess their importance.

\section{Construction of Bateman data set}\label{appC}

Bateman data set is based on the resolution of the Bateman equations, which is an ODE system modeling multi species reactions: 

\begin{equation*}
    \partial_t\boldsymbol{\eta}(t) = \boldsymbol{\Sigma_r}\big(\boldsymbol{\eta}(t)\big) \cdot \boldsymbol{\eta}(t)
\text{,    with initial conditions    }
    \boldsymbol{\eta}(0) = \boldsymbol{\eta_0}, 
\end{equation*}
and $\boldsymbol{\eta}\in(\mathbb{R}^{+})^M$, $\boldsymbol{\Sigma}_r \in\mathbb{R}^{M\times M}$. Here, $f: ( \boldsymbol{\eta_0}, t) \rightarrow \boldsymbol{\eta}(t)$, and we are interested in $\boldsymbol{\eta}(t)$, which is the concentration of each of the species $S_k$, with $k \in \{1,...,M\}$. For physical applications, $M$ ranges from tens to thousands. We consider the particular case $M=11$. Matrix $\boldsymbol{\Sigma_r}\big(\boldsymbol{\eta}(t)\big)$ depends on reaction constants. Here, $4$ reactions are considered and each reaction $p$ has constant $\sigma_p$.

\begin{equation*}
    \begin{dcases}
        (1): S_1 + S_2 \rightarrow S_3 + S_4 + S_6 + S_7,  \\
        (2): S_3 + S_4 \rightarrow S_2 + S_8 + S_{11},  \\
        (3): S_2 + S_{11} \rightarrow S_3 + S_5 + S_9, \\
        (4): S_3 + S_{11} \rightarrow S_2 + S_5 + S_6 + S_{10},
     \end{dcases}
\end{equation*}
with $\sigma_1=1$, $\sigma_2=5$, $\sigma_3=3$ and $\sigma_4=0.1$. To obtain $\boldsymbol{\Sigma_r}\big(\boldsymbol{\eta}(t)\big)$, the species have to be considered one by one. Here we give an example of how to construct the second row of $\boldsymbol{\Sigma_r}\big(\boldsymbol{\eta}(t)\big)$. The other rows are built the same way. Given the reaction equations : 
\begin{equation*}
    \partial_t \eta_2 = - \sigma_1\eta_1\eta_2 + \sigma_2\eta_3\eta_4 - \sigma_3\eta_2\eta_{11} + \sigma_4\eta_3\eta_{11},
\end{equation*}
because $S_2$ disappears in reactions (1) and (3) involving $S_1$ and $S_{11}$ as other reactants at rate $\sigma_1$ and $\sigma_3$, respectively, and appears in reactions (2) and (4) involving $S_3$, $S_4$ and $S_3$, $S_{11}$ as reactants, at rate $\sigma_2$ and $\sigma_4$ respectively. Hence, the second row of $\boldsymbol{\Sigma_r}\big(\boldsymbol{\eta}(t)\big)$ is 
\begin{equation*}
    [0, \; -\sigma_1 \eta_1, \; 0, \; \sigma_2 \eta_3,  \; 0, \; 0, \; 0, \; 0, \; 0, \; 0, \; -\sigma_3 \eta_2 + \sigma_4 \eta_3 ],
\end{equation*}
with $\boldsymbol{\eta}(t)$ denoted by $\boldsymbol{\eta}$ to simplify the equation and $\eta_i$ the $i$-th component of $\boldsymbol{\eta}$.
To construct the training, validation and test data sets, we sample uniformly $( \boldsymbol{\eta}_0, t) \in [0,1]^{12} \times [0,5]$ $130000$ times. We denote these samples $( \boldsymbol{\eta}_0, t)_i $ for $i \in \{1,...,130000\}$. Then, we apply a first order Euler solver with a time step of $10^{-3}$ to compute $f(( \boldsymbol{\eta}_0, t)_i)$. As a result, neural network's input is $( \boldsymbol{\eta}_0, t)$ and neural network's output is $f(( \boldsymbol{\eta}_0, t))$.
\end{appendices}

\end{document}